\documentclass[letterpaper]{article} 
\usepackage{aaai2026}  
\usepackage{times}  
\usepackage{helvet}  
\usepackage{courier}  
\usepackage[hyphens]{url}  
\usepackage{graphicx} 
\usepackage{natbib}  
\usepackage{caption} 
\usepackage{algorithm}
\usepackage{dsfont}
\usepackage{booktabs}
\usepackage{multicol}
\usepackage{multirow}
\usepackage{enumitem}
\usepackage{xcolor}
\usepackage{array}
\usepackage{makecell}
\usepackage[most]{tcolorbox}
\usepackage{url}
\usepackage{amsmath,amssymb}
\usepackage{amsfonts}
\usepackage{algpseudocode}
\usepackage{adjustbox} 
\usepackage{newfloat}
\usepackage{listings}
\DeclareCaptionStyle{ruled}{labelfont=normalfont,labelsep=colon,strut=off} 
\floatstyle{ruled}
\newfloat{listing}{tb}{lst}{}
\floatname{listing}{Listing}
\usepackage{newfloat}
\usepackage{listings}
\DeclareCaptionStyle{ruled}{labelfont=normalfont,labelsep=colon,strut=off} 
\floatstyle{ruled}
\newfloat{listing}{tb}{lst}{}
\floatname{listing}{Listing}
\title{Forecasting Clinical Risk from Textual Time Series: Structuring Narratives for Temporal AI in Healthcare}

\author{
    Shahriar Noroozizadeh\textsuperscript{\rm 1, 2}\equalcontrib,
    Sayantan Kumar\textsuperscript{\rm 3}\equalcontrib,
    Jeremy C. Weiss\textsuperscript{\rm 3}
}

\affiliations{
    \textsuperscript{\rm 1}Machine Learning Department, School of Computer Science, Carnegie Mellon University, Pittsburgh, PA, USA\\
    \textsuperscript{\rm 2} Heinz College of Information Systems and Public Policy, Carnegie Mellon University, Pittsburgh, PA, USA\\
    \textsuperscript{\rm 3} National Library of Medicine, National Institutes of Health, Bethesda, MD, USA\\
}

\begin{document}

\maketitle

\begin{abstract}
Clinical case reports encode temporal patient trajectories that are often underexploited by traditional machine learning methods relying on structured data. 
In this work, we introduce the forecasting problem from textual time series, where timestamped clinical findings---extracted via an LLM-assisted annotation pipeline---serve as the primary input for prediction. 
We systematically evaluate a diverse suite of models, including fine-tuned decoder-based large language models and encoder-based transformers, on tasks of event occurrence prediction, temporal ordering, and survival analysis. 
Our experiments reveal that encoder-based models consistently achieve higher F1 scores and superior temporal concordance for short- and long-horizon event forecasting, while fine-tuned masking approaches enhance ranking performance. 
In contrast, instruction-tuned decoder models demonstrate a relative advantage in survival analysis, especially in early prognosis settings. 
Our sensitivity analyses further demonstrate the importance of time ordering, which requires clinical time series construction, as compared to text ordering, the format of the text inputs that LLMs are classically trained on. 
This highlights the additional benefit that can be ascertained from time-ordered corpora, with implications for temporal tasks in the era of widespread LLM use. 
\end{abstract}

%

\section{Introduction}

Healthcare disparities persist globally, with preventable deaths from conditions like sepsis disproportionately affecting underserved populations who often present to under-resourced facilities with limited access to specialized expertise. In these settings, much of the critical diagnostic and prognostic information exists only in unstructured clinical narratives—case reports, discharge summaries, and progress notes—as comprehensive structured data infrastructure may be unavailable~\cite{anzalone2025lower,seinen2025using}. While machine learning approaches have shown promise on structured data, with recent work demonstrating that incorporating text alongside structured inputs significantly improves predictive performance~\citep{kline2022multimodal}, large language models struggle with clinical risk estimation, especially in patient-facing scenarios~\citep{wongposition}.
This gap highlights the need for automated risk forecasting from textual clinical records that could democratize access to expert-level clinical reasoning, particularly in resource-constrained environments where timely specialist consultation is not available.
The challenge of LLM risk estimation motivates a structured treatment of risk forecasting and survival modeling from unstructured narrative sources.

Among textual sources, retrospective case reports serve as a rich training ground for developing temporally-aware AI systems that can eventually be deployed on real-time clinical notes. Case reports provide holistic accounts of clinical trajectories with explicit temporal reasoning—exactly the type of structured thinking needed for real-time clinical decision support. However, their narrative structure interleaves temporally unordered information, making them difficult to apply in forecasting tasks. Without identifying the time of occurrence for each event, models are prone to causal leakage—using information not available at the time of prediction. Solving this temporal reasoning challenge in case reports establishes the foundation for deployment in live clinical environments where such capabilities could improve patient outcomes.

\begin{figure*}[!tbp] 
    \centering
    \includegraphics[width=.95\linewidth]{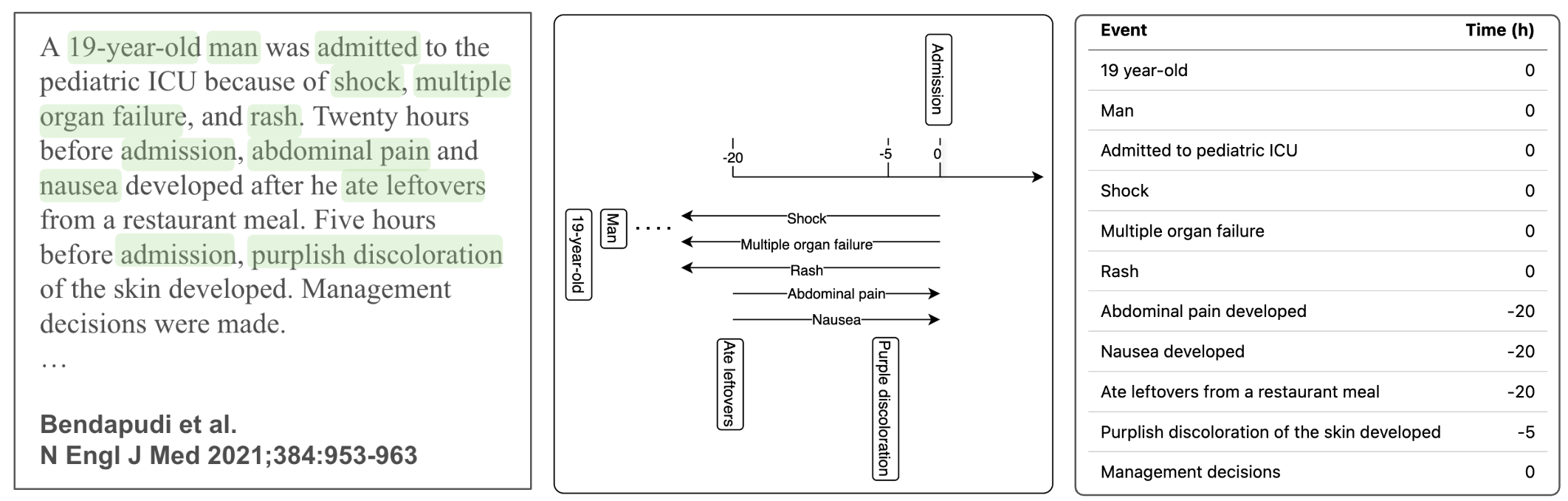} 
    \caption{Example case report (left) with timeline representations (middle) and text-ordered, (event, time) tuples (right).}
    \label{fig:annotation_ex}
\end{figure*}

One might ask whether large pretrained language models can already perform forecasting from clinical narratives, given their exposure to biomedical texts~\citep{peng2020empirical,weber2024redpajama}. However, these models have architectural limitations for temporal reasoning tasks. Encoder-based models apply random masking that rarely yields temporally coherent representations, while decoder-style models mask text in linear sequence, modeling events in text order rather than time order. 
Our experiments make these limitations explicit. For instance, on the timeline in Figure \ref{fig:annotation_ex}, the chance of recovering a valid time-ordered mask is below $0.002\%$, and decoder models show low temporal concordance (e.g., $c=0.23$). Without a task-specific restructuring, such factors lead to suboptimal forecasting performance.

To address these temporal reasoning challenges, we develop a forecasting framework that
transforms free-text narratives into textual time series---sequences of (event, time) tuples---and makes explicit the temporal structure of patient trajectories. Building on recent annotation work~\citep{wang2025large,noroozizadeh2025pmoa}, we propose a hybrid pipeline combining rule-based heuristics and LLM-assisted extraction to recover temporally localized clinical findings, validated on real clinical notes with expert verification.

Our work builds on extensive research in clinical NLP and temporal reasoning~\citep{kohane1987temporal, leeuwenberg2020towards, cheng2023typed}, but shifts the emphasis from extraction to downstream clinical forecasting. While prior efforts have focused on entity recognition and temporal relation extraction \citep{zhou2007temporal, uzuner20112010, sun2013evaluating}, few have systematically assessed how such representations influence downstream tasks such as risk prediction and survival modeling. We bridge this gap by formalizing textual time series representations and evaluating their effectiveness for temporally grounded clinical prediction.

We empirically assess this framework across multiple tasks and modeling strategies, benchmarking encoder and decoder models—both fine-tuned and prompted versions—on three forecasting tasks: event prediction, survival analysis, and temporal ordering of future clinical events. Our results show no single model dominates across all tasks, highlighting the importance of aligning model architecture with forecasting objectives. We further examine how input ordering (time vs. narrative) impacts generalization and analyze the sensitivity of models to missing historical context through systematic masking experiments.

In summary, this paper makes the following contributions:

(i) \textbf{Annotation and Extraction Pipeline:} Building on prior methods for extracting information from the PubMed Open Access corpus \cite{noroozizadeh2025reconstructing}, this work introduces a pipeline that transforms case reports into textual time series via regular expression filtering and LLM-assisted extraction, yielding temporally anchored (event, time) tuples while minimizing causal leakage by restricting input to past events;
\mbox{(ii) \textbf{Comprehensive Model Comparison:}} 
    We conduct a comprehensive evaluation of encoder- and instruction-tuned decoder-based models—using both fine-tuned MLP heads and prompting—across event prediction, temporal ordering, and survival analysis tasks. Results show no single model excels universally, underscoring the need to align model choice with the specific forecasting objective;
\mbox{(iii) \textbf{Temporal- versus Text- Ordering:}}
    We investigate the impact of annotation order by comparing training on time-ordered versus text-ordered data. Results show that preserving the narrative's natural order can improve generalization to external datasets, while time-based ordering can enhance ranking performance;
\mbox{(iv) \textbf{Sensitivity Analysis of Temporal Masking:}}
    We conduct systematic dropout experiments by randomly masking parts of the clinical history to assess their impact on forecasting and event ordering. While higher masking levels reduce classification performance (F1), the concordance index remains largely stable, highlighting differing sensitivities of binary prediction and ranking tasks to historical context in textual time series;
(v) \textbf{Methodological Framework for Temporal Clinical Forecasting:} Our approach demonstrates how narrative clinical texts can be systematically converted into structured temporal representations for forecasting tasks, providing a replicable methodology that could be adapted to other clinical and non-clinical text sources.

\section{Related Work}

\paragraph{Temporal Information Extraction from Clinical Text:}
Extracting timelines from clinical narratives is a challenging biomedical NLP task. The i2b2 2012 challenge introduced annotated datasets for temporal relation extraction from discharge summaries \citep{uzuner20112010}. Subsequent methods linked clinical events to timestamps or temporal expressions \citep{leeuwenberg2020towards, frattallone2024using}, typically assuming pre-defined event spans. We adopt our prior approach \citep{noroozizadeh2025reconstructing, noroozizadeh2025pmoa} to assign timepoints to findings from full-length case reports, enabling finer temporal resolution. Unlike methods using structured EHR data, we focus solely on text—crucial for sources like PubMed that lack structured metadata. By directly supervising event-time alignment, we overcome limitations of span-based annotations~\citep{rosenbloom2011data}. This aligns with growing emphasis on temporality in sepsis and critical care phenotyping \citep{johnson2018comparative, kamran2024evaluation}, particularly given high missingness rates in structured data \citep{johnson2023mimic,seinen2025using}.

\paragraph{Predictive Modeling with Clinical Text:}
Clinical text has been used for outcome prediction tasks like mortality and readmission, with models like \textit{ClinicalBERT} achieving high levels of performance on EHR notes \citep{huang2019clinicalbert, gu2021domain}. 
Text captures complementary information—symptoms and social factors—not found in structured codes. 
However, traditional approaches using full-text or bag-of-words representations obscure temporal dynamics. 
Our approach preserves event order within narratives, treating data as time series. This aligns with recent efforts adapting LLMs for time series tasks, such as Time-LLM mapping numeric sequences to tokens \citep{jin2023time}. 
To our knowledge, this is the first work to forecast from LLM-derived time series in clinical domains. 
While prior sepsis prediction models rely on vitals or scores like SOFA and SAPS \citep{hou2020predicting, noroozizadeh2023temporal}, our method is complementary—leveraging narrative descriptions that include clinician interpretations and context. 
This is valuable when case reports or patient histories are available but structured data is sparse or unavailable.

\paragraph{Large Language Models in Healthcare:}
Recent LLMs such as \texttt{GPT-3} and \texttt{GPT-4} have enabled applications from medical QA to note summarization. These models encode substantial medical knowledge and reasoning capacity—\texttt{GPT-4} demonstrates strong performance on board exams and can support patient record abstraction for quality reporting \citep{boussina2024large}. However, their use in clinical forecasting remains largely unexplored.
Our benchmark tests LLMs via zero- and few-shot prompting to assess clinical event forecasting from patient narratives, comparing results to fine-tuned models to quantify the gap between general-purpose and task-specific modeling. This contributes evidence that while LLMs hold broad medical knowledge, fine-tuning and structured inputs are often required for clinical reliability \citep{huang2019clinicalbert, gu2021domain}. Some studies suggest domain-specific tuning does not always improve over foundation models \citep{jeong2024medical}, reinforcing the need for task-specific evaluations like ours.

\section{Methods\footnote{Our code can be found at: \url{https://github.com/Shahriarnz14/Textual-Time-Series-Forecasting}.\\
\hspace*{1.8em}The extended version of this paper, including the full technical appendix, is available in \citet{noroozizadeh2025forecasting}.}}

\begin{figure}
    \centering
    \includegraphics[width=\linewidth]{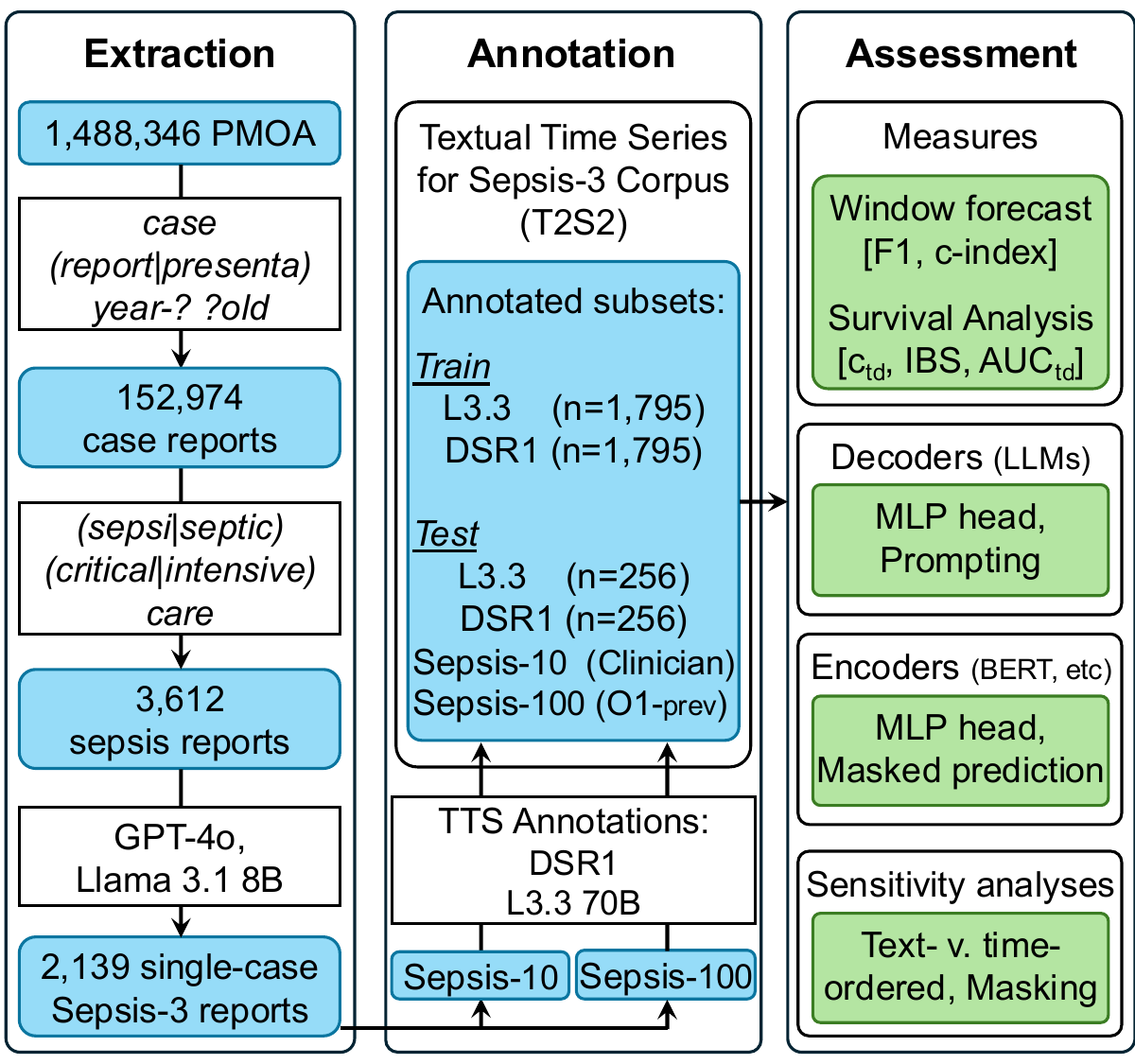}
    \caption{\textit{T2S2} forecasting analysis pipeline}
    \label{fig:flowchart}
\end{figure}

We next describe our dataset construction and annotation pipeline, define the forecasting and survival prediction tasks, detail the modeling approaches evaluated, and present the sensitivity analyses used to assess robustness (Figure~\ref{fig:flowchart}). 

\subsection{Dataset Extraction}
We used the PubMed Open Access (PMOA) Subset (as of April 25, 2024) for our analysis. Following prior work \cite{wang2025large}, we extracted the body of each case by selecting the text between the \texttt{==== Body} and \texttt{==== Ref} delimiters in the PMOA corpus. To maintain relevance, we included only documents with case-insensitive matches to \texttt{(case report|case presenta)} and \texttt{year-? ?old}, an approach shown to be more specific than relying on PubMed metadata \cite{noroozizadeh2025reconstructing}.

Additional filtering was applied to identify potential sepsis-related case reports using the regular expressions \texttt{sepsi|\allowbreak septic} and \texttt{critical|intensive} care. We then used LLM-based queries to extract key attributes, including sepsis diagnosis, patient count, age, and gender. For this step, both \texttt{GPT-4o} and \texttt{Llama-3.1-8B-Instruct} were employed. Reports describing more than one case were excluded. A case report was retained if either model classified it as involving a Sepsis-3 diagnosis.

This multi-step process yielded a total of 2319 Sepsis-3 case reports. From this set, we randomly selected two subsets for evaluation: a group of 10 reports (\textit{sepsis-10}), which underwent expert clinical annotation to serve as a gold standard, and a larger group of 100 reports (\textit{sepsis-100}) as a bronze standard for broader testing based on an alternative annotator (\texttt{O1-preview}; Figure \ref{fig:flowchart}). 

\subsection{Textual Time Series Annotation}
\label{sec:tts-annotation}

Following \citeauthor{noroozizadeh2025reconstructing} \citeyearpar{noroozizadeh2025reconstructing}, a textual time series case report corpus of Sepsis-3 patients was constructed from the PMOA sepsis reports. A textual time series refers to a sequence of clinical findings, each paired with a timestamp (either absolute or relative to the case presentation time), corresponding to an individual patient. Here, a clinical finding is defined as a free-text expression describing an entity that pertains to or may impact a person's health.

A textual time series differs from conventional approaches to clinical concept annotation (e.g., \cite{sun2013evaluating,uzuner20112010}) in its extension of the text span to better capture the specificity and contextual meaning of each clinical finding. In our work, the interpretation of a clinical finding differs from the i2b2 concept guidelines in the following respects: (1) Clinical findings are not restricted to single prepositional phrases following a markable entity. For instance, instead of splitting “pain in chest that radiates substernally” into less informative parts, we retain the full phrase as a single, complete finding to preserve its meaning and context. (2) To improve clarity, compound phrases are broken into individual findings. For example, “metastases in the liver and pancreas” becomes “metastasis in the liver” and “metastasis in the pancreas.”

We used \texttt{DeepSeek-R1-UD-IQ1} and \texttt{Llama-3.3-70B-Instruct} models to generate clinical textual time series from the 2319 sepsis case reports. We refer to this as \textbf{T}extual \textbf{T}ime-\textbf{S}eries for \textbf{S}epsis (\textit{T2S2}).
Details on the quality evaluation of LLM-extracted temporal annotations and the prompts used for extraction are provided in Appendix~\ref{apd:tts_validity} and~\ref{TTS_prompt}, respectively.
Our dataset was split into training (n = 1,795; 80\% train, 20\% validation), testing \textit{T2S2}-test (n = 244), and two external validation sets: \textit{sepsis-10} (n = 10) and \textit{sepsis-100} (n = 90).

\subsection{Forecasting Tasks}
Using the timelines extracted from the annotations, we defined two primary forecasting tasks and a survival analysis task as follows:

\paragraph{Event Occurrence Prediction.} 
Given a prefix of the clinical timeline (all events up to a certain time $t$), the model is tasked with predicting whether each of the immediate next $k$ events occurs within a specified time horizon. 
This setup is repeated across multiple time cut-offs to simulate an “online” forecasting scenario, where the model must output a binary label for each of the next $k$ events:
does this event occur within $h$ hours after $t$? 
Time horizons used include 1 hour, 24 hours (1 day), and 168 hours (1 week). 
The task is framed as a series of binary classification problems, with one binary decision per event. 
Evaluation is based on precision, recall, and F1 score, averaged across event positions for each time horizon.

\paragraph{Temporal Ordering Prediction.} 
This task assesses the model's ability to reconstruct the correct sequence of future events. 
At each time cut-off $t$, we extract the next $k$ events from the timeline and remove their timestamps. 
The model must output a permutation of these events that matches their true chronological order.
This is framed as a ranking task, evaluated by computing the pairwise concordance between the predicted and true orderings (e.g., proportion of correctly ordered pairs). 
This tests whether the model can infer temporal progression from unordered event content alone.

\paragraph{Survival Analysis for Mortality Time.}  
We include a classical survival analysis task to model time until death. 
Many case reports specify whether the patient died and, if so, when (e.g., "the patient died on hospital day 10"), which enables us to define a time-to-event outcome. 
For this task, we evaluate models at predefined cut-off times--- specifically at 0 hours (admission), 24 hours (1 day), and 168 hours (1 week)--- and use the extracted event history up to each cut-off as input. 
A survival model is trained to predict the probability of survival over time beyond each cut-off. 
We evaluate using the time-dependent concordance to measure alignment of the predicted survival times with actual outcomes.

\subsection{Modeling Approaches}
To evaluate performance on the event forecasting and survival prediction tasks, we implement five modeling paradigms: (i) decoder-only large language models (LLMs) with fine-tuned heads, (ii) prompted LLMs without gradient updates (zero- or few-shot), (iii) encoder-only models with task-specific fine-tuned heads, (iv) encoder-masking models with fine-tuning, and (v) encoder-masking models in zero-shot settings. In Appendix~\ref{apd:encoder-forecasting} and~\ref{apd:encoder-survival}, we provide technical details of encoder-based forecasting methods and the survival modeling framework respectively.

\paragraph{(i) Fine-tuned LLMs.}  
We apply instruction-tuned decoder-only models from the \textit{Llama} and \textit{DeepSeek} families: \textit{Llama-3.3-70B-Instruct}, \textit{Llama-3.1-8B-Instruct}, and their distilled variants \citep{grattafiori2024llama, liu2024deepseek}. We also include open-source models with documented training corpora (\textit{OLMO-32B-Instruct} \citep{olmo20242}, \textit{RedPajama-INCITE-7B-Instruct} \citep{weber2024redpajama}) to address potential data leakage concerns---notably, RedPajama explicitly excludes PubMed from its training data. Additionally, we evaluate a medically fine-tuned decoder model (\textit{MediPhi-PubMed} \citep{corbeil2025modular}) to assess domain-specific benefits.

Each model is paired with a lightweight multilayer perceptron (MLP) head trained for classification or ranking, depending on the task. 
The input consists of a text-formatted prefix of events (e.g., a clinical timeline up to time $t$), optionally accompanied by an instruction template. 
The output layer produces task-specific predictions: binary labels for event occurrence or a permutation over $k$ events for ordering. 
Training is conducted using cross-entropy loss for classification and pairwise ranking loss for ordering.

\paragraph{(ii) Prompted LLMs.}  
In the zero- or few-shot setting, we use the same LLM architectures as above, but without any fine-tuning. 
Instead, we supply structured prompts at inference time to guide the model toward task-specific outputs. 
Each prompt includes: 
(1) a system instruction establishing the model’s role (e.g., "You are an expert physician."), 
(2) a user instruction describing the prediction task, and 
(3) one or more input-output examples in a few-shot format.
Prompt designs are customized for each task and are detailed in Appendix~\ref{apd:prompts}. Model outputs are parsed to extract binary labels or ordered lists, and errors in generation or ambiguity are excluded from evaluation.

\paragraph{(iii) Fine-tuned Encoder-Only Models.}  
We also evaluate a range of encoder-based models trained end-to-end on each task. 
Architectures include general-purpose models (\textit{BERT-base-uncased} \citep{devlin2019bert}, \textit{RoBERTa-base} \citep{liu2019roberta}, \textit{DeBERTa-v3-small} \citep{he2021debertav3}, \textit{ModernBERT-base/large} \citep{warner2024smarter}) and biomedically-pretrained variants (\textit{BioClinical-ModernBERT-base/large} \citep{sounack2025bioclinical}).
For each model, we append a task-specific MLP head for event occurrence or ordering prediction. 
The model input is the same flattened prefix of the event sequence used in other settings, tokenized and formatted according to the respective architecture. 
These models are trained using standard supervised learning objectives and evaluated on the same metrics as other methods: F1 score for event occurrence, and pairwise concordance for temporal ordering. 
For survival analysis, the final hidden states are passed into a time-to-event prediction head and evaluated using the time-dependent concordance index \citep{antolini2005time}.

\paragraph{(iv, v) Encoder-Masking Models.}
We adapt masked language modeling (MLM) for our temporal reasoning tasks. For event occurrence, models predict masked tokens (``\texttt{yes}''/``\texttt{no}'') in prompts such as ``Will [event] happen within 24 hours? \texttt{[MASK]}''. For temporal ordering, models predict ``\texttt{before}''/``\texttt{after}'' tokens in prompts comparing event pairs. We evaluate both fine-tuned variants (trained on task-specific objectives) and zero-shot variants (using pretrained MLM capabilities without additional training). This approach leverages the bidirectional context of encoder models while maintaining interpretable predictions through constrained vocabulary.

\subsection{Sensitivity Analyses}
We conduct sensitivity analyses to examine how the temporal structure and completeness of input sequences affect forecasting performance. These analyses probe the robustness of models to changes in event ordering and to varying levels of missing historical information.

\paragraph{Time-ordering Strategies.} We evaluate two strategies for ordering clinical events within textual time series inputs. In the \emph{text-ordered} setting, events are presented in the order they appear in the original case report narratives, preserving the narrative structure produced during extraction. In the \emph{time-ordered} setting, events are sorted chronologically by their documented occurrence time, enforcing strict temporal alignment. This comparison isolates the effect of narrative sequencing versus explicit temporal ordering on model performance. For text-ordered inputs, to prevent causal leakage, events occurring after the forecast cutoff time $t$ are masked for encoder-based models and omitted for decoder models.

\paragraph{Timestep Dropout.} To assess robustness to incomplete patient histories, we introduce a \emph{timestep dropout rate} (TDR), defined as the proportion of input timesteps randomly removed prior to inference. This procedure simulates varying degrees of missing clinical documentation that may arise in deployment settings. We vary TDR from 0\% (full history) to 90\% (severely truncated history) in increments, masking events independently and uniformly at random. This setup enables controlled evaluation of model performance under partial information, probing the extent to which predictive accuracy and event ordering depend on the completeness of historical context.

\begin{table*}[!htbp]
\centering
\resizebox{\linewidth}{!}
{
\begin{tabular}{lcccccccc|cccccccc}
\toprule
& \multicolumn{8}{c|}{\textbf{\textit{T2S2} (\texttt{DeepSeek-R1})}} & \multicolumn{8}{c}{\textbf{\textit{T2S2} (\texttt{Llama-3.3-70B})}} \\
& F1(1h) & F1(1d) & F1(1w) & c-index & F1(1d)-10 & c10 & F1(1d)-100 & c100 & F1(1h) & F1(1d) & F1(1w) & c-index & F1(1d)-10 & c10 & F1(1d)-100 & c100 \\
\hline
\textbf{LLM-MLP head} &&&&&&&&&&&&&&&&\\
DS-L3.3 70B & \textbf{0.075} & 0.482 & \textbf{0.796} & \textbf{0.632} & \textbf{0.397} & \textbf{0.578} & \textbf{0.613} & 0.595 & 0.140 & 0.563 & 0.811 & \textbf{0.624} & \textbf{0.397} & \textbf{0.585} & \textbf{0.629} & \textbf{0.598} \\
OLMO 32B Instruct & 0.000 & 0.411 & 0.688 & 0.621 & 0.273 & 0.561 & 0.432 & 0.593 & 0.190 & 0.315 & 0.764 & 0.604 & 0.238 & 0.551 & 0.315 & 0.596 \\
RedPajama-INCITE 7B Instruct & 0.000 & 0.352 & 0.651 & 0.618 & 0.282 & 0.572 & 0.471 & 0.594 \\
MediPhi-PubMed & 0.000 & 0.268 & 0.751 & 0.626 & 0.169 & 0.561 & 0.210 & 0.597 & 0.000 & 0.507 & 0.801 & 0.623 & 0.277 & 0.564 & 0.534 & 0.600 \\ 
\textbf{Prompting} &&&&&&&&&&&&&&&&\\
L3.3 70B few-shot & \textbf{0.095} & \textbf{0.380} & 0.652 & -- & \textbf{0.360} & -- & 0.452 & -- & 0.064 & \textbf{0.460} & 0.729 & -- & 0.313 & -- & 0.480 & -- \\
OLMO 32B few-shot ordinal & 0.013 & 0.315 & 0.651 & -- & 0.247 & -- & 0.422 & -- & 0.046 & 0.413 & 0.647 & -- & 0.300 & -- & 0.424 & -- \\
\hline
\textbf{Encoder-MLP head} &&&&&&&&&&&&&&&&\\
ModernBERT-base & 0.306 & \textbf{0.576} & 0.877 & 0.646 & 0.428 & 0.562 & 0.607 & 0.558 & 0.257 & 0.645 & \textbf{0.903} & 0.594 & 0.431 & 0.590 & 0.626 & 0.565 \\
Bioclinical ModernBERT-base & 0.264 & 0.559 & \textbf{0.879} & \textbf{0.677} & 0.395 & \textbf{0.598} & \textbf{0.635} & \textbf{0.610} & 0.290 & \textbf{0.653} & 0.902 & \textbf{0.650}
& \textbf{0.449} & \textbf{0.618} & \textbf{0.662} & \textbf{0.627} \\ 
\hline
\textbf{Encoder-masking-fine-tuned} &&&&&&&&&&&&&&&&\\
ModernBERT-base & \textbf{0.186} & 0.449 & \textbf{0.697} & 0.672 & 0.325 & 0.576 & \textbf{0.503} & 0.613 & \textbf{0.139} & \textbf{0.489} & 0.700 & 0.632 & \textbf{0.334} & 0.595 & \textbf{0.478} & 0.612 \\
Bioclinical ModernBERT-base & 0.166 & 0.450 & 0.692 & \textbf{0.676} & 0.336 & 0.604 & 0.499 & \textbf{0.655} & 0.129 & 0.481 & 0.733 & 0.653 & 0.331 & \textbf{0.615} & 0.475 & \textbf{0.658} \\ 
\hline
\textbf{Encoder-masking-zeroshot} &&&&&&&&&&&&&&&&\\
ModernBERT-base & 0.169 & \textbf{0.468} & \textbf{0.804} & 0.496 & \textbf{0.358} & 0.501 & \textbf{0.511} & 0.499 & 0.132 & \textbf{0.525} & \textbf{0.840} & 0.556 & \textbf{0.352} & 0.503 & 0.511 & 0.501 \\
Bioclinical ModernBERT-base & \textbf{0.246} & 0.458 & 0.614 & \textbf{0.498} & 0.301 & 0.509 & 0.457 & 0.498 & \textbf{0.197} & 0.460 & 0.533 & 0.499 & 0.299 & 0.520 & 0.465 & 0.499 \\
\bottomrule
\end{tabular}
}
\caption{Forecasting performance (event occurrence: F1 and correct event ordering: concordance-index) of the ensuing $k=8$ events. All models are trained/fine-tuned on time-ordered annotations from either \texttt{DeepSeek-R1} or \texttt{Llama 3.3-70B}. \textbf{Bold} values indicate best in category within each column group (refer to Tables \ref{tab:DS_forecasting} and \ref{tab:L33_forecasting} in Appendix \ref{apd:full_tables} for detailed performance statistics on all models). Performance statistics for all models are presented in \ref{tab:DS_forecasting} and \ref{tab:L33_forecasting}, respectively. Abbreviations: F1(1d)-10/100: F1 (1 day) for sepsis-10 and sepsis-100 respectively. c10/100: c-index for sepsis-10 and sepsis-100 respectively.
}
\label{tab:forecasting_clean_combined}
\end{table*}


\section{Results}

In this section, we present results for our three main evaluation settings: event forecast within a subsequent time window, temporal ordering of forecasted events, and survival analysis. We provide sensitivity analyses on the forecasting tasks, examining the effects of temporal ordering and historical context availability in Appendix~\ref{apd:sensitivity_analyses}~and~\ref{apd:masking_l33_dsr1}.

\subsection{Forecasting Tasks}
\paragraph{Event forecast within next 24 hours: F1 performance.}

Our results show that encoder-based models outperform decoder-based LLMs in event forecasting across both \texttt{DeepSeek-R1} and \texttt{Llama-3.3-70B} annotations (Table~\ref{tab:forecasting_clean_combined}).
Forecasting performance across all models is shown in Tables \ref{tab:DS_forecasting} and \ref{tab:L33_forecasting} (Appendix~\ref{apd:full_tables}).
Encoder models, especially those with a fine-tuned MLP head, achieve substantially higher F1 scores than decoder LLMs, reinforcing the effectiveness of encoder-based representations for forecasting.

Among decoders, we observe varied performance patterns. \textit{RedPajama-INCITE-7B-Instruct}, whose training excludes PubMed, achieves 0.352 F1 at 24h---better than some decoder models like \textit{MediPhi-PubMed} (0.268) and within the typical decoder performance range, though below top performers like \textit{DeepSeek-Llama-3.3-70B} (0.482). \textit{OLMO-32B-Instruct} performs competitively (0.411), suggesting that open models without biomedical pretraining can still achieve strong temporal forecasting results.

Among encoder models, both general-purpose and biomedically-pretrained variants show strong performance. While \textit{ModernBERT-base} achieves slightly higher F1 scores on internal test sets (0.576 vs 0.559), \textit{BioClinical-ModernBERT-base} demonstrates superior concordance (0.677 vs 0.646 for \textit{DeepSeek-R1}) and notably better generalization to external datasets, with F1 scores of 0.635 and 0.662 on \textit{sepsis-100} compared to 0.607 and 0.626 for the general-purpose variant. This suggests that while both \textit{ModernBERT} architectures are effective, biomedical pretraining particularly enhances temporal reasoning and cross-dataset generalization.

Different training strategies yield distinct performance patterns. The MLP head fine-tuning approach consistently outperforms both fine-tuned masking and zero-shot masking models, particularly in long-horizon forecasting. While \textit{ModernBERT-base} achieves strong F1 scores, \textit{BioClinical-ModernBERT-base} demonstrates the best overall performance when considering both accuracy and generalization. Zero-shot masking models, particularly standard \textit{BERT} and \textit{RoBERTa}, fail to make meaningful predictions at the 1-hour mark. However, \textit{ModernBERT} variants exhibit relatively better zero-shot performance, with \textit{BioClinical-ModernBERT-base} achieving notably higher scores in zero-shot settings (0.246 F1 at 1h), indicating that biomedical pretraining improves out-of-the-box temporal reasoning.

Performance trends remain consistent across forecasting windows. F1 scores improve as the window increases, with 1-hour predictions being most challenging and 168-hour predictions yielding highest scores. Models perform worse on external validation sets than internal \textit{T2S2} test sets, though \textit{BioClinical-ModernBERT} models show the smallest performance degradation, maintaining their advantage in cross-dataset generalization.

\paragraph{Temporal ordering of forecasted events: concordance.}

Encoder models consistently outperform decoders in correctly ranking the order of upcoming events, as measured by concordance (c-index; Table \ref{tab:forecasting_clean_combined}; complete results in Tables~\ref{tab:DS_forecasting} and~\ref{tab:L33_forecasting} for \texttt{DeepSeek-R1} and \texttt{Llama-3.3-70B} annotations, respectively, with full details in Appendix~\ref{apd:full_tables}. Decoder models with potential biomedical exposure achieve modest concordance (0.618-0.632), while \textit{RedPajama-INCITE-7B-Instruct} without PubMed data in its training corpus performs similarly, suggesting domain-specific knowledge is less critical here than for event prediction. The comparable performance of \textit{RedPajama-INCITE-7B-Instruct} to other general models (with access to PubMed in their pretraining) also provides evidence that the \textit{T2S2} prediction task is distinct from raw PubMed text and assuages causal leakage concerns.

Biomedical pretraining in encoders demonstrates clear advantages, with \textit{BioClinical-ModernBERT-base} achieving the highest concordance (0.677 with an MLP head, 0.676 with fine-tuned masking) and strong performance across datasets. MLP-head training yields superior F1 scores, while fine-tuned masking excels in concordance. \textit{BioClinical-ModernBERT-base} also generalizes well to external datasets (c-index \textgreater0.60), suggesting that biomedical pretraining provides stability in temporal reasoning under dataset shifts.

\paragraph{Survival analysis: model performance across timepoints and cohorts.}
Tables~\ref{tab:concordance_t2s2_only} and~\ref{tab:concordance_wide} (Appendix~\ref{apd:full_survival}) show time-dependent concordance results where 
instruction-tuned decoder models clearly outperform encoder-based models. This pattern contrasts with the forecasting tasks (Table~\ref{tab:forecasting_clean_combined}), where encoders excelled. Across both the \textit{T2S2} and \textit{sepsis-100} cohorts, decoder models consistently achieve the strongest results. Notably, \textit{RedPajama-INCITE-7B-Instruct} attains the highest concordance overall (0.76 at 168h, \textit{sepsis-100}) despite lacking any biomedical text in pretraining, while larger models such as \textit{DeepSeek-R1-Llama-70B} and \textit{OLMO-32B-Instruct} also perform robustly. Encoder models---including biomedically fine-tuned variants like \textit{BioClinical-ModernBERT-base}, which is competitive on \textit{sepsis-100} but not on \textit{T2S2}---are generally surpassed by decoder models.

Predictive performance does not universally improve with more observation time. Models evaluated on the \textit{T2S2} cohort, often peak at earlier windows (0h or 24h) rather than at 168h. Meanwhile, the \textit{sepsis-10} cohort experiences clear ceiling effects, with many models achieving perfect concordance (1.00). These results suggest that, unlike the forecasting tasks, survival prediction in our dataset benefits more from the capabilities inherent in large decoder models than from domain-specific pretrained encoders.

\begin{table}[!tbp]
\centering
{
\begin{tabular}{lccc}
\toprule
\textbf{Model} & 0h & 24h & 168h \\
\midrule
bert-base-uncased       & 0.60 & 0.61 & 0.53 \\
roberta-base            & 0.55 & 0.60 & 0.60 \\
deberta-v3-small        & 0.56 & 0.57 & 0.57 \\
ModernBERT-base         & 0.52 & 0.58 & 0.58 \\
ModernBERT-large        & 0.53 & 0.59 & 0.58 \\
\textcolor{black}{Bioclinical ModernBERT-base} & 0.57 & 0.53 & 0.53 \\
\textcolor{black}{Bioclinical ModernBERT-large} & 0.54 & 0.55 & 0.54 \\
DeepSeek-R1-Llama-70B   & \textbf{0.64} & \textbf{0.63} & 0.59 \\
Llama-3.3-70B-Instruct  & 0.62 & \textbf{0.63} & 0.58 \\
DeepSeek-R1-Llama-8B    & 0.60 & 0.58 & 0.58 \\
Llama-3.1-8B-Instruct   & 0.62 & 0.61 & \textbf{0.61} \\
\textcolor{black}{OLMO 32B Instruct} & 0.60 & 0.62 & 0.59 \\
\textcolor{black}{RedPajama-INCITE 7B Instruct} & 0.56 & 0.60 & \textbf{0.61} \\
\textcolor{black}{MediPhi-PubMed} & 0.54 & 0.62 & 0.55 \\
\bottomrule
\end{tabular}
}
\caption{
Time-dependent concordance index for survival analysis on the \texttt{DeepSeek-R1} \textit{T2S2} annotations evaluated at 0h, 24h, and 168h. Extended \textit{sepsis-10} and \textit{sepsis-100} results are provided in Appendix~\ref{apd:full_survival}.
}
\label{tab:concordance_t2s2_only}
\end{table}

\section{Discussion and Conclusion}

Our findings highlight the superiority of encoder-based models over decoder-based LLMs for event forecasting, emphasizing the limitations of autoregressive models in structured prediction tasks. Among encoder models, \textit{BioClinical-ModernBERT-base} consistently performs best, achieving superior concordance (0.677 vs 0.646) and better external generalization despite slightly lower internal F1 scores than general-purpose \textit{ModernBERT-base}. MLP-head fine-tuning excels in F1 scores while fine-tuned masking models achieve higher concordance, particularly in long-term forecasting. This suggests that classification and ranking tasks benefit from distinct optimization strategies.

The consistent advantage of biomedically-pretrained models across tasks deserves special attention. While general-purpose models achieve strong internal performance, biomedical pretraining provides crucial benefits for real-world deployment: \textit{BioClinical-ModernBERT} variants show the smallest performance degradation on external datasets and maintain higher zero-shot capabilities (0.246 F1 at 1h vs near-zero for standard \textit{BERT}/\textit{RoBERTa}). This suggests that domain-specific pretraining enhances not just accuracy but also robustness and generalization—critical factors when considering clinical deployment where distribution shifts are common.

The gradual improvement of F1 scores over longer time horizons indicates that event patterns become more predictable over time, whereas short-term forecasting remains challenging due to higher variability. Performance drops on external validation datasets highlight generalization challenges, though \textit{ModernBERT} models exhibit relative robustness. For survival analysis, instruction-tuned LLMs like \textit{Llama-3.3-70B-Instruct} outperform traditional transformer baselines, with several models achieving peak concordance at early timepoints rather than with extended observation.

Our sensitivity analysis reveals several trade-offs: time-ordered training generally improves concordance while text-ordering can yield better F1 scores on externally-annotated datasets. Robustness experiments show F1 scores degrade significantly beyond 60\% timestep dropout, while concordance remains stable, indicating event ranking is less sensitive to partial history than event classification. 

Additional insights into the forecasting results are presented in Appendix~\ref{apd:additional_insights}. Specifically, we summarize two observations that clarify model behavior and apparent anomalies: (i) why decoders fare better on survival but not short-horizon forecasting, and (ii) interpreting very low $F_{1}$ scores and external validation drops.

\textbf{Methodological Contributions and Potential Impact.} Our framework demonstrates how narrative clinical texts can be systematically converted into structured temporal representations for forecasting tasks. The ability to extract temporally structured insights from unstructured clinical text could potentially support clinical decision-making, particularly in settings with limited structured data or specialized expertise. 

Importantly, while consumers frequently consult LLMs about health risk via prompting in chats, this study demonstrates that the prompting approach performs substantially worse in risk prediction than alternative approaches, at least with respect to precision/recall/F1, with prompted LLMs achieving at best 0.460 F1 at 24h compared to 0.653 for fine-tuned encoders.
Our study highlights several alternative approaches and characterizes their relative performance strengths and weaknesses.
In addition, our finding that concordance remains stable with 60\% missing context suggests robustness to incomplete documentation scenarios, which highlights the degradation pattern in performance as it relates to context availability. 

Our approach also demonstrates that our language model systems can reliably capture the temporal reasoning that clinicians use in practice based on their clinical documentation. However, additional work would be required to validate performance on real-time clinical data and demonstrate measurable impact on patient outcomes in deployment. Further discussion on societal impact of our work is in Appendix \ref{apd:broader_impact}.

\textbf{Limitations and Future Directions.} 
Our study has important limitations to consider. First, our pipeline relies on case reports from the PubMed Open Access (PMOA) corpus. Because these reports often highlight rare or atypical presentations and differ from routine clinical notes (e.g., progress notes, discharge summaries), generalizability to real-world health-system text may be limited. However, case reports provide rich, temporally structured narratives with explicit clinical rationale, making them well suited for evaluating model performance under sparse documentation. Accordingly, we present a general framework for textual time series forecasting that uses case reports as an interpretable, temporally rich foundation for method development, with extensions to real-world corpora such as MIMIC discharge summaries. See Appendix~\ref{apd:beyond_pmoa} for a detailed discussion on the generalizability of our pipeline beyond case report details, as well as comparisons with other clinical sources.

Second, while we focused on sepsis due to its clinical relevance and prevalence in case reports, our framework is fundamentally disease-agnostic. Preliminary results from over 125K PMOA case reports across various diagnostic conditions show promising generalization, with future work planned on broader diagnostic categories and real-world corpora such as MIMIC-IV discharge summaries.

Finally, despite mitigation steps (timestamp extraction, temporal masking, external test sets), pretrained language models may still have been exposed to PMOA content. Although they were not trained on our derived \emph{(event, time)} sequences or temporally reordered data, the original narratives could appear in pretraining. Our pipeline applies temporal reordering and masking of future events beyond the prediction cutoff, enforcing a causal structure absent from the originals and yielding a distinct setup that requires reasoning over partial, temporally grounded inputs. 
While we included \textit{RedPajama-INCITE-7B-Instruct}, an open-source model with a documented PubMed-free corpus for pretraining, and saw its performance (0.352 F1 at 24h; concordance 0.618) was comparable to decoders potentially exposed to PubMed, other strategies supporting the argument that the \textit{T2S2} task is sufficiently distinct from raw PMOA text and mitigates causal leakage concerns could be examined.
See Appendix~\ref{apd:open_source} for a detailed discussion of how open-source models mitigate potential data leakage, as well as additional forecasting results on 115 case reports published after the pretraining cutoffs of encoder-based models (post-2020).



\section*{Acknowledgments}
This research was supported in part by the Division of Intramural Research (DIR) of the National Library of Medicine (NLM), National Institutes of Health. This work utilized the computational resources of the NIH HPC Biowulf cluster. S.N. was supported by Carnegie Mellon University TCS Presidential Fellowship, and Natural Sciences and Engineering Research Council of Canada (NSERC) PGS-D award. S.N. was also supported in part by an appointment to the National Library of Medicine Research Participation Program administered by the Oak Ridge Institute for Science and Education (ORISE) through an interagency agreement between the U.S. Department of Energy (DOE) and the National Library of Medicine, National Institutes of Health. ORISE is managed by ORAU under DOE contract number DE-SC0014664. All opinions expressed in this paper are the authors’ and do not necessarily reflect the policies and views of NIH, NLM, DOE, or ORAU/ORISE.
\bibliography{aaai2026}

\appendix
\setcounter{secnumdepth}{1}

\renewcommand{\thetable}{A\arabic{table}}
\setcounter{table}{0}

\begin{center}
    {\LARGE \bfseries Supplementary Material}
\end{center}
\vspace{1em} 

This supplementary material provides detailed information organized as follows:
\begin{itemize}
    \item \textbf{Appendix~\ref{apd:tts_validity}}: Validity of LLM-extracted temporal annotations (adopted from our prior work (adopted from our prior work \cite{wang2025large}).
    \item \textbf{Appendix~\ref{TTS_prompt}}: LLM prompt template for extracting clinical time series from case reports.
    \item \textbf{Appendix~\ref{apd:prompts}}: Prompting strategies for zero-shot and few-shot forecasting.
    \item \textbf{Appendix~\ref{apd:encoder-forecasting}}: Technical details of encoder-based forecasting methods, including MLP-head and masking approaches.
    \item \textbf{Appendix~\ref{apd:encoder-survival}}: Survival modeling framework using language model embeddings and the hyperparameters used.
    \item \textbf{Appendix~\ref{apd:full_tables}}: Complete forecasting performance tables for all models on DeepSeek-R1 and Llama-3.3-70B annotations.
    \item \textbf{Appendix~\ref{apd:full_survival}}: Complete survival analysis performance table for all datasets.
    \item \textbf{Appendix~\ref{apd:sensitivity_analyses}}: Sensitivity analysis of the forecasting performance when using time-ordered versus text-ordered training.
    \item \textbf{Appendix~\ref{apd:masking_l33_dsr1}}: Sensitivity analysis of masking history on Llama-3.3-70B and DeepSeek-R1 annotations.
    \item \textbf{Appendix~\ref{apd:additional_insights}}: Additional insights into forecasting results.
    \item \textbf{Appendix~\ref{apd:open_source}}: Analysis of open-source models to address data leakage concerns.
    \item \textbf{Appendix~\ref{apd:beyond_pmoa}}: Generalizability beyond PubMed case reports: comparison with other clinical sources.
    \item \textbf{Appendix~\ref{apd:broader_impact}}: Broader societal impact and ethical considerations.
\end{itemize}

\section{Validity of Temporal Annotations}
\label{apd:tts_validity}


We used \texttt{DeepSeek-R1-UD-IQ1} and \texttt{Llama-3.3-70B-Instruct} models to generate clinical textual time series from
the 2319 sepsis case reports. We refer to this as Textual
Time Series for Sepsis (\textit{T2S2}). We assessed the quality of \textit{T2S2} using three key metrics: event match rate, temporal concordance (c-index), and log-time discrepancy on two external validation sets: \textit{sepsis-10} (n = 10) and \textit{sepsis-100} (n = 90).(adopted from our prior work \cite{wang2025large}).

\subsection*{Text Event Match}
To align predicted clinical findings with reference annotations, we use a recursive best-match procedure adapted from \citep{wang2025large}. At each step, we select the reference-prediction pair with the minimal string-based distance; ties are broken by text order (\emph{i.e.}, the order of events in the annotation files). Matched items are removed and the process recurses on the remainder, preventing many-to-one mappings. We evaluated Levenshtein distance, cosine similarity of \textit{BERT} (\textit{bert-base-uncased}) embeddings, and sentence-embedding cosine similarity from \textit{PubMedBERT} (\textit{S-PubMedBert-MS-MARCO}). Prior validation showed \textit{PubMedBERT-based} cosine similarity with a threshold of 0.1 performed best, and we used this metric for evaluation. Pseudocode is provided in Algorithm \ref{alg:recursive_match}.

\subsection*{Temporal Ordering}
Temporal ordering accuracy was quantified with the concordance index (c-index), the probability that a randomly chosen pair of events is correctly ordered. For reference times $\{t^r_i\}$ and predicted times $\{t^p_i\}$,
\[
\text{c-index}=\frac{1}{N}\sum_{i<j}\mathds{1}\{(t^r_i-t^r_j)(t^p_i-t^p_j)>0\},
\]
where $N$ is the number of comparable pairs (event indices $i,j$ with $t^r_i\neq t^r_j$ and $t^p_i\neq t^p_j$), and $\mathds{1}\{\cdot\}$ is the indicator function. Higher values indicate better agreement with the reference timeline.

\begin{algorithm}[tbp]
\caption{Recursive Best Match}
\label{alg:recursive_match}
\begin{algorithmic}[1]
\Require Two lists: \texttt{ref} (reference events) and \texttt{pred} (predicted events)
\Ensure List of best-matching event pairs
\Function{MatchEvents}{\texttt{ref}, \texttt{pred}}
    \If{\texttt{ref} is empty \textbf{or} \texttt{pred} is empty}
        \State \Return $[]$
    \EndIf
    \State $\text{min\_distance} \gets \infty$
    \State $\text{best\_pair} \gets \text{None}$
    \ForAll{$r \in \texttt{ref}$}
        \ForAll{$p \in \texttt{pred}$}
            \State $d \gets \text{ComputeDistance}(r, p)$
            \If{$d < \text{min\_distance}$}
                \State $\text{min\_distance} \gets d$
                \State $\text{best\_pair} \gets (r, p)$
            \ElsIf{$d = \text{min\_distance}$}
                \State Get indices of $r$, $p$, and current $\text{best\_pair}$
                \If{index of $r <$ best $r$ index}
                    \State $\text{best\_pair} \gets (r, p)$
                \ElsIf{index of $r = $ best $r$ index \textbf{and} index of $p <$ best $p$ index}
                    \State $\text{best\_pair} \gets (r, p)$
                \EndIf
            \EndIf
        \EndFor
    \EndFor
    \State Remove $\text{best\_pair.r}$ from \texttt{ref}
    \State Remove $\text{best\_pair.p}$ from \texttt{pred}
    \State $\text{result} \gets [\text{best\_pair}] + \text{MatchEvents}(\texttt{ref}, \texttt{pred})$
    \State \Return $\text{result}$
\EndFunction
\end{algorithmic}
\end{algorithm}

\subsection*{Time Discrepancy Analysis}
 Temporal prediction accuracy was assessed via discrepancies between predicted and reference timestamps. We computed the log-time discrepancy as $\log(1+\Delta t)$, where $\Delta t$ is the absolute time difference (in hours), and summarized errors with the Area Under the Log-Time Cumulative Distribution Function (AULTC), which reflects how rapidly the error CDF rises—larger AULTC indicates better temporal alignment. 

\begin{table}[thbp]
\centering
\begin{adjustbox}{max width=\columnwidth}
\setlength{\tabcolsep}{4.5pt} 
\renewcommand{\arraystretch}{1.1}

\begin{tabular}{lccc}
\hline
\multicolumn{4}{c}{\textbf{\textit{sepsis-10}}} \\
\hline
\textbf{Model} & \textbf{Event Match Rate} & \textbf{Concordance} & \textbf{AULTC} \\
\texttt{DeepSeek-R1}      & 0.79 & 0.876 & 0.772 \\
\texttt{Llama-3.3-70B}    & 0.75 & 0.932 & 0.759 \\
\hline
\multicolumn{4}{c}{\textbf{\textit{sepsis-100}}} \\
\hline
\textbf{Model} & \textbf{Event Match Rate} & \textbf{Concordance} & \textbf{AULTC} \\
\texttt{DeepSeek-R1}      & 0.91 & 0.875 & 0.801 \\
\texttt{Llama-3.3-70B}    & 0.78 & 0.894 & 0.791 \\
\hline
\end{tabular}
\end{adjustbox}
\caption{Comparison of event match rate (Event), median concordance (c), median absolute error (MAE), and area under the log-time curve (AULTC) across \texttt{DeepSeek-R1} and \texttt{Llama-3.3-70B} for \textit{sepsis-10}, and \textit{sepsis-100} datasets.}
\label{tab:sepsis_onecol}
\end{table}

Formally, we compute the AULTC as the area under $F(x)$ from $x = 0$ to $x = \log(1 + S_{\text{max}})$, normalized by $\log(1 + S_{\text{max}})$:

\begin{equation*}
\label{eq:aultc}
\begin{aligned}
\text{AULTC}
&= \frac{1}{\log(1+S_{\max})}
   \sum_{i=1}^{k}\big(x_{(i)}-x_{(i-1)}\big)\frac{i}{k} \\
&\quad + \frac{\log(1+S_{\max})-x_{(k)}}{\log(1+S_{\max})},
\end{aligned}
\end{equation*}

where $x_{(0)} = 0$. With this definition, AULTC = 1 indicates that discrepancies are zero (perfect recovery), resulting in maximum area $\log(1 + S_{\text{max}})$, and AULTC = 0 indicates that all discrepancies exceed $S_{\text{max}}$, yielding zero area. We also stratified results by reference time intervals (hour, day, week, year) to examine accuracy as a function of time from case presentation.

\subsection*{Annotation Quality Results}
On the \textit{sepsis-10} subset, \texttt{DeepSeek-R1} achieved an event match rate of 0.79, c-index of 0.876, and AULTC of 0.772, while \texttt{Llama-3.3-70B} obtained 0.75, 0.932, and 0.759, respectively. Thus, \texttt{Llama} provided stronger pairwise ordering (higher c-index) but with fewer matched events and a slightly weaker log-time error profile (lower AULTC) than \texttt{DeepSeek}. On \textit{sepsis-100}, \texttt{DeepSeek-R1} reached 0.91 (event match rate), 0.875 (c-index), and 0.801 (AULTC), whereas \texttt{Llama-3.3-70B} reached 0.78, 0.894, and 0.791. Here, \texttt{DeepSeek} matched substantially more events and achieved better AULTC, while \texttt{Llama} again showed a small c-index advantage. Overall, \texttt{DeepSeek} tends to optimize event identification and log-time alignment, whereas \texttt{Llama} slightly favors pairwise temporal ordering. 

Overall, these metrics together indicate strong alignment of the generated textual time series with expert annotations, sufficient to support forecasting benchmarks.

\onecolumn

{\refstepcounter{section} \label{TTS_prompt}
\Large\bfseries\thesection\quad LLM Prompt to Generate Clinical Textual Time Series from PMOA Case Reports
}

\vspace{1em}

\begin{minipage}[!htbp]{0.99\textwidth}
\begin{tcolorbox}[colback=gray!5, colframe=gray!100, title=Prompt, boxrule=2pt]

{You are a physician. 

Extract the clinical events and the related time stamp from the case report. The admission event has timestamp 0. If the event is not available, we treat the event, e.g. current main clinical diagnosis or treatment with timestamp 0. The events that happened before event with 0 timestamp have negative time, the ones after the event with 0 timestamp have positive time. The timestamps are in hours. The unit will be omitted when output the result. If there is no temporal information of the event, please use your knowledge and events with temporal expression before and after the events to provide an approximation. We want to predict the future events given the events happened in history. For example, here is the case report.}\\

\noindent
\textcolor{teal}{\texttt{An 18-year-old male was admitted to the hospital with a 3-day history of fever and rash. Four weeks ago, he was diagnosed with acne and received the treatment with minocycline, 100 mg daily, for 3 weeks. With increased WBC count, eosinophilia, and systemic involvement, this patient was diagnosed with DRESS syndrome. The fever and rash persisted through admission, and diffuse erythematous or maculopapular eruption with pruritus was present. One day later the patient was discharged.}}\\

\noindent
{Let's find the locations of event in the case report, it shows that four weeks ago of fever and rash, four weeks ago, he was diagnosed with acne and receive treatment. So the event of fever and rash happened four weeks ago, 672 hours, it is before admitted to the hospital, so the time stamp is -672. diffuse erythematous or maculopapular eruption with pruritus was documented on the admission exam, so the timestamp is 0 hours, since it happens right at admission. DRESS syndrome has no specific time, but it should happen soon after admission to the hospital, so we use our clinical judgment to give the diagnosis of DRESS syndrome the timestamp 0. then the output should look like:}\\
\vspace{2mm}
\noindent 

\texttt{18 years old | 0 \\
male | 0 \\
admitted to the hospital | 0 \\
fever | -72 \\
rash | -72 \\
acne | -672 \\
minocycline | -672 \\
increased WBC count | 0 \\
eosinophilia | 0 \\
systemic involvement | 0 \\
diffuse erythematous or maculopapular eruption | 0 \\
pruritus | 0 \\
DRESS syndrome | 0 \\
fever persisted | 0 \\
rash persisted | 0 \\
discharged | 24}\\

\noindent {Separate conjunctive phrases into its component events and assign them the same timestamp (for example, separation of ‘fever and rash’ into 2 events: ‘fever’ and ‘rash’). If the event has duration, assign the event time as the start of the time interval. Attempt to use the text span without modifications except ‘history of’ where applicable. Include all patient events, even if they appear in the discussion; do not omit any events; include termination/discontinuation events; include the pertinent negative findings, like ‘no shortness of breath’ and ‘denies chest pain’. Show the events and timestamps in rows, each row has two columns: one column for the event, the other column for the timestamp. The time is a numeric value in hour unit. The two columns are separated by a pipe $\mid$ as a bar-separated file. Skip the title of the table. Reply with the table only.}

\end{tcolorbox}
\end{minipage}

\twocolumn

\onecolumn

{\refstepcounter{section}
\noindent\Large\bfseries\thesection\quad Prompting Strategies for Forecasting Tasks
\label{apd:prompts}}

\subsection{Zero-Shot}

\begin{tcolorbox}[colback=gray!5, colframe=gray!100, title=Prompt, boxrule=2pt]

{You are an expert physician.}

{Reply to the prompt with structured predictions in a k-item, bar-separated row.

For example, if there are k=3 events (A, B, C) and only B occurs in the time window, then:}
{0 $\mid$ 1 $\mid$  0
would be the correct response. \\ 

Reply with the answer only. Do NOT provide any other text. Do NOT explain.}

\end{tcolorbox}

\subsection{Zero-Shot (with degree of confidence)}

\begin{tcolorbox}[colback=gray!5, colframe=gray!100, title=Prompt, boxrule=2pt]
{You are an expert physician.}

{Reply to the prompt with structured predictions in a k-item, bar-separated row.
Assess whether the event happens in the time window using a scale from 1 to 5, where 5 means it definitely happened and 1 means it did not happen.

For example, if there are k=3 events (A, B, C) and only B occurs in the time window, then:
$1 | 5 | 1$
would be the correct response. \\

Reply with the answer only. Do NOT provide any other text. Do NOT explain.}
\end{tcolorbox}

\subsection{Few-Shot}
\begin{tcolorbox}[colback=gray!5, colframe=gray!100, title=Prompt, boxrule=2pt]

{You are an expert physician.}

{Reply to the prompt with predictions in a k-item, bar-separated row to answer if the k events happen within a given forecast time window. \\

Example 1. If there are k=3 events (A, B, C) and only B occurs in the requested time window, then:
0 $\mid$ 1 $\mid$ 0
would be the correct response.}\\

{Example 2. Suppose the time series is:}

\noindent \texttt{\textcolor{teal}
{admitted | 0 \\
history of diabetes | 0 \\
elevated troponin | 5 \\
--- \textbf{PREDICTION WINDOW} --- \\
cardiac catheterization | 6 \\
admission to icu | 24 \\ 
transfer to floor | 72\\ 
discharge | 144\\ 
office visit | 1440}
}

{Then, if the prediction is at 5 hours, the time preceding ``--- PREDICTION WINDOW ---'', the events admitted, history of diabetes, and elevated troponin have occurred.  Supposing k=4 and a forecast window of 24 hours, then the task it to determine which of cardiac catheterization, admission to icu, transfer to floor, and discharge happen. So the correct response for the example is:
1 $\mid$ 1 $\mid$ 0 $\mid$ 0
because cardiac catherization (6) and admission to icu (24) happen before 5 + 24 = 29.}\\

{Reply with the answer only. Do NOT provide any other text. Do NOT explain.}

\end{tcolorbox}

\subsection{Few-Shot Ordinal}

\begin{tcolorbox}[colback=gray!5, colframe=gray!100, title=Prompt, boxrule=2pt]

{You are an expert physician.}

{Reply to the prompt with predictions in a k-item, bar-separated row to answer if the k events happen within a given forecast time window, using a scale from 1 to 5, where 5 means it definitely happened and 1 means it did not happen.}\\

{Example 1. If there are k=3 events (A, B, C) and only B occurs in the time window, then:
1 $\mid$ 5 $\mid$ 1
would be the correct response.}\\

{Example 2. Suppose the time series is:}\\
\noindent \texttt{\textcolor{teal}{
admitted | 0\\
history of diabetes | 0\\
elevated troponin | 5\\
--- \textbf{PREDICTION WINDOW} ---\\
cardiac catheterization | 6\\
admission to icu | 24\\
transfer to floor | 72\\
discharge | 144\\
office visit | 1440}}

{Then, if the prediction is at 5 hours, the time preceding ``--- PREDICTION WINDOW ---'', the events admitted, history of diabetes, and elevated troponin have occurred.  Supposing k=4 and a forecast window of 24 hours, then the task it to determine which of cardiac catheterization, admission to icu, transfer to floor and discharge happen. So the an excellent response for the example is:
5 $\mid$ 4 $\mid$ 1 $\mid$ 1
because cardiac catherization (6) and admission to icu (24) happen before 5 + 24 = 29.}\\

{Reply with the answer only. Do NOT provide any other text. Do NOT explain.}

\end{tcolorbox}

\twocolumn


\clearpage
\section{Textual Time Series Forecasting Framework of Encoder Models}
\label{apd:encoder-forecasting}

We develop a general framework for predictive modeling over textual time series of the encoder-only models, designed to accommodate the irregular, event-based structure common to clinical case reports. 
As highlighted Section~\ref{sec:tts-annotation}, each data instance consists of a sequence of clinical events, where each event is a free-text string describing a clinical finding, intervention, or diagnosis, paired with a timestamp in hours relative to the time of admission. 
The format resembles a two-column table, where the first column contains textual clinical events and the second column contains integer-valued timestamps.

\paragraph{Input Preprocessing and Timeseries Construction.}
From each case report, we construct multiple training examples for sequence forecasting. 
For a fixed forecast horizon parameter $K$, we generate overlapping windows of historical events and forecast targets. 
Specifically, for each case report with $L$ rows, we construct a sequence of up to $L-K$ forecasting examples. 
Each example is structured as follows:

\begin{itemize}
    \item A \textbf{history segment} $\mathcal{H}_t = \{(e_i, t_i)\}_{i=1}^{n}$, consisting of clinical events $e_i$ occurring at timestamps $t_i \leq t$.
    \item A \textbf{forecast target} $\mathcal{F}_t = \{(f_j, \tau_j)\}_{j=1}^{K}$, consisting of the next $K$ events after $t$.
\end{itemize}

To preserve the temporal coherence of events, we treat all events with the same timestamp as an atomic unit. 
That is, if multiple events share the same timestamp, they are included together in the history or forecast block, and the next forecasting window is only advanced after the full timestamp block has been consumed.

\paragraph{Model Input Representation.}
Each training example is serialized into a transformer-compatible input by linearizing the historical events and forecast events. 
Importantly, we preserve both temporal and textual content by encoding each event in the form ``\texttt{t:e}'' where \texttt{t} is the timestamp and \texttt{e} is the event string.
The input to the language model takes the form:

\begin{equation}
\texttt{[CLS]} \; t_1\texttt{:}e_1 \; \texttt{[SEP]} \; t_2\texttt{:}e_2 \; \texttt{[SEP]} \; \dots \; \texttt{[SEP]} \; t_n\texttt{:}e_n
\end{equation}

This is followed by additional target-specific components, as described below.

\paragraph{Language Models.}
To test the generality of our framework, we experiment with a variety of pre-trained encoder-only language models, including:
\begin{itemize}
    \item \textit{bert-base-uncased} (BERT; Devlin et al., 2018)
    \item \textit{roberta-base} (Liu et al., 2019)
    \item \textit{microsoft/deberta-v3-small} (He et al., 2021)
    \item \textit{answerdotai/ModernBERT-base}, \textit{answerdotai/ModernBERT-large}
    \item \textit{thomas-sounack/BioClinical-ModernBERT-base}, \textit{thomas-sounack/BioClinical-ModernBERT-large} 
\end{itemize}
Each model is used with its corresponding tokenizer. 
For models with SentencePiece tokenization (e.g., \textit{DeBERTa}), we use the appropriate fast tokenizer such as \textit{DebertaV2Tokenizer} to ensure compatibility. 
All inputs are truncated from the left (i.e., keeping the most recent tokens) if they exceed the model's context window (typically 512 tokens).

\paragraph{Prediction Tasks.}
We define two primary prediction tasks over $\mathcal{F}_t$, each formulated as a binary classification problem.

\subparagraph{1. Event Ordering (Concordance) Task.}
The goal in this task is to predict the correct temporal ordering between future events. 
For each example, we generate all \emph{comparable pairs} of events in $\mathcal{F}_t$, defined as pairs $(f_i, f_j)$ where $\tau_i \neq \tau_j$. 
For each pair, we form a binary classification example as follows:

\begin{itemize}
    \item \textbf{Input:} The serialized history segment followed by:
    \begin{equation}
        \texttt{[SEP]} \; \tau_i\texttt{:}f_i \; \texttt{[SEP]} \; \tau_j\texttt{:}f_j \; \texttt{[SEP]}
    \end{equation}
    \item \textbf{Label:} $y = 1$ if $\tau_i < \tau_j$, else $y = 0$
\end{itemize}

The model output is taken from the \texttt{[CLS]} token, passed through a linear classification head, and softmaxed to produce a binary probability. 
We optimize cross-entropy loss across all pairwise comparisons. 
At evaluation time, we report the concordance index (c-index), which measures the proportion of correctly predicted pairs among all comparable pairs.

\subparagraph{2. Time-Window Classification Task.}
This task evaluates whether each forecast event occurs within a fixed prediction window $H$ hours of the last historical event. 
For each $f_j \in \mathcal{F}_t$, we generate an input:

\begin{itemize}
    \item \textbf{Input:} The serialized history segment followed by:
    \begin{equation}
        \texttt{[SEP]} \; \tau_j\texttt{:}f_j \; \texttt{[SEP]}
    \end{equation}
    \item \textbf{Label:} $y = 1$ if $\tau_j - t_n \leq H$, where $t_n$ is the timestamp of the last historical event.
\end{itemize}

This is again treated as a binary classification task over the \texttt{[CLS]} embedding, using a linear layer and softmax. 
We report macro-averaged F1 scores over all prediction examples.

\paragraph{Masked Language Modeling Approaches.}
In addition to the MLP-head approach described above, we implement masked language modeling (MLM) variants that leverage the pretrained masked token prediction capabilities of encoder models. These approaches reformulate our forecasting tasks as cloze-style problems:

\subparagraph{1. Masked Event Occurrence Prediction.}
For the time-window classification task, we construct prompts that include a \texttt{[MASK]} token whose prediction indicates whether an event occurs within the specified horizon:

\begin{itemize}
    \item \textbf{Input:} History formatted as: \texttt{[CLS]} $h_1$ \texttt{[SEP]} $h_2$ \texttt{[SEP]} ... \texttt{[SEP]} Will "$e_j$" happen within $H$ hours? \texttt{[MASK]} \texttt{[SEP]}
    \item \textbf{Target:} The model predicts either "\texttt{yes}" or "\texttt{no}" at the \texttt{[MASK]} position
\end{itemize}

We constrain the output vocabulary to only these two tokens and compute cross-entropy loss over their logits. For fine-tuned variants, we train the entire model to optimize this masked prediction task. For zero-shot variants, we directly use the pretrained model's predictions without additional training.

\subparagraph{2. Masked Temporal Ordering.}
For the concordance task, we predict relational tokens that indicate temporal ordering:

\begin{itemize}
    \item \textbf{Input:} History followed by: \texttt{[SEP]} $e_1$ \texttt{[MASK]} $e_2$ \texttt{[SEP]}
    \item \textbf{Target:} The model predicts "\texttt{before}" if $e_1$ occurs before $e_2$, else "\texttt{after}"
\end{itemize}

Again, we restrict predictions to these two tokens. This formulation allows the model to leverage its pretrained understanding of temporal relationships encoded during MLM pretraining. 

Note that in our dataset, we randomize the relative ordering of $e_1$ and $e_2$ each time to avoid introducing positional biases that could lead the model to rely on superficial cues rather than genuine temporal reasoning.

\subparagraph{Implementation Details.}
For both tasks, we extract the logits at the \texttt{[MASK]} position and apply a linear projection to the target vocabulary size (2 tokens). The key advantage of this approach is that it can leverage pretrained masked language modeling capabilities, potentially requiring less task-specific training. We evaluate both fine-tuned versions (which update all model parameters) and zero-shot versions (which use frozen pretrained weights).

\paragraph{Training and Evaluation.}
Each task is trained separately using AdamW with linear learning rate decay and early stopping based on validation loss. 
We save the model checkpoint with the best validation performance. 
During inference, we reload this checkpoint and evaluate on a held-out test set (which is obtained from textual time series of unseen case reports).

For both tasks, we extract softmax probabilities rather than hard labels to allow for threshold calibration and ROC-based analyses. 
All training is conducted using PyTorch and HuggingFace Transformers.
\clearpage
\section{Survival Modeling with Language Model Embeddings}
\label{apd:encoder-survival}

\begin{table}[b]
\centering
\begin{adjustbox}{max width=\columnwidth}
\begin{tabular}{ll}
\toprule
\textbf{Model} & \textbf{Hyperparameters} \\
\midrule
RSF & \texttt{n\_estimators} $\in$ \{20,50,100,200,500,1000\} \\
    & \texttt{min\_samples\_split} $\in$ \{2,10,20,50,100\} \\
    & \texttt{min\_samples\_leaf} $\in$ \{1,5,10,20\} \\
DeepSurv/DeepHit & \texttt{num\_nodes} $\in$ \{32,64,128,256,512,768,1024,1200,\\
& 1536,1800,2048,3072,4096\} \\
    & \texttt{dropout} $\in$ \{0.1,0.2,0.4,0.5\} \\
\bottomrule
\end{tabular}
\end{adjustbox}
\caption{Hyperparameter grid search ranges for our survival models.}
\label{tab:hyperparams}
\end{table}

To evaluate the prognostic value of textual information encoded in large language models (LLMs), we adopt a two-stage framework: (1) extraction of fixed-dimensional sequence embeddings from various pre-trained LLMs, and (2) downstream survival modeling using these embeddings as covariates.

In the first stage, we process each textual time series (which is converted to a textual context of \emph{"(time) clinical event [SEP] ... [SEP] (time) clinical event [SEP]"}) using a suite of LLMs to obtain dense vector representations that summarize the content of the input sequence. 
For models belonging to the encoder family, including \textit{bert-base-uncased}, \textit{roberta-base}, \textit{deberta-v3-small}, \textit{ModernBERT-base}, and \textit{ModernBERT-large}, we extract the final hidden state corresponding to the \textit{[CLS]} token, which is conventionally used to represent the entire sequence in classification tasks. 
This token-specific embedding serves as a compact, sequence-level representation.

For decoder-based models that do not utilize a dedicated \texttt{[CLS]} token, including \textit{DeepSeek-R1-Distill-Llama-70B}, \textit{Llama-3.3-70B-Instruct}, \textit{DeepSeek-R1-Distill-Llama-8B}, and \textit{Llama-3.1-8B-Instruct}, we compute the mean-pooled embedding over the last hidden states of all non-padding tokens. 
This pooling strategy yields a fixed-length vector that captures the overall semantic content of the input while mitigating the impact of padding artifacts.

In the second stage, these embedding vectors are used as input covariates to three survival models, each designed to capture time-to-event dynamics in different ways:

\begin{itemize}
    \item \textbf{Random Survival Forest (RSF):} 
    A nonparametric ensemble method based on decision trees, capable of handling complex interactions and high-dimensional inputs. 
    The model is tuned over the number of trees (\texttt{n\_estimators}), the minimum number of samples required to split an internal node (\texttt{min\_samples\_split}), and the minimum number of samples at a leaf node (\texttt{min\_samples\_leaf}); see Table~\ref{tab:hyperparams} for search ranges.

    \item \textbf{DeepSurv:} 
    A neural-network generalization of the Cox proportional hazards model. 
    The architecture comprises fully connected layers with varying widths (\texttt{num\_nodes}) and dropout rates; see Table~\ref{tab:hyperparams} for search ranges. The number of training epochs is fixed.

    \item \textbf{DeepHit:} 
    A multi-task neural network that jointly models the discrete hazard and survival functions. 
    As with DeepSurv, we perform a grid search over hidden layer sizes and dropout rates while keeping the number of epochs fixed; see Table~\ref{tab:hyperparams}.
\end{itemize}

Hyperparameter tuning for each survival model is performed using a validation split from the training set. 
For each combination of language model and survival model, we conduct grid search over the respective hyperparameter space, selecting the configuration that maximizes the time-dependent concordance index \citep{antolini2005time} on the validation set. 
Final evaluation is conducted on held-out test sets (\textit{T2S2\_test}, \textit{sepsis-10}, \textit{sepsis-100}), and results are aggregated across survival metrics to assess model performance.

This framework allows for a systematic evaluation of how different pre-trained LLMs, and their associated embedding strategies, contribute to downstream survival prediction tasks.
\clearpage
\onecolumn

{\refstepcounter{section}
\noindent\Large\bfseries\thesection\quad Forecasting Performance on \texttt{DeepSeek-R1} and \texttt{Llama-3.3} Textual Time Series Annotations
\label{apd:full_tables}}

\vspace{0em}

Tables~\ref{tab:DS_forecasting} and~\ref{tab:L33_forecasting} present the forecasting performance of our models on textual time series annotations generated by \texttt{DeepSeek-R1} and \texttt{Llama-3.3-70B}, respectively. We report standard forecasting metrics across multiple evaluation settings to provide a comprehensive view of model performance. 

\begin{table*}[!htbp]
\footnotesize
\centering
\resizebox{\linewidth}{!}{
\begin{tabular}{lcccccccc}
\toprule
& \multicolumn{4}{c}{\textit{T2S2} (\texttt{DeepSeek-R1})} & \multicolumn{2}{c}{\textit{sepsis-10}} & \multicolumn{2}{c}{\textit{sepsis-100}} \\
\cmidrule(lr){2-5} \cmidrule(lr){6-7} \cmidrule(lr){8-9}
& F1(1h) & F1(1d) & F1(1w) & c-index & F1(1d) & c-index & F1(1d) & c-index \\
\midrule
\textbf{LLM-MLP head} &&&&&&&&\\
DS-L3.3 70B & \textbf{0.075} & 0.482 & \textbf{0.796} & \textbf{0.632} & \textbf{0.397} & \textbf{0.578} & \textbf{0.613} & 0.595 \\
L3.3 70B & 0.000 & 0.295 & 0.764 & 0.631 & 0.312 & 0.569 & 0.404 & 0.597 \\
DS-L3.1 8B & 0.000 & \textbf{0.492} & 0.650 & 0.630 & 0.381 & 0.566 & 0.601 & 0.600 \\
L3.1 8B & 0.000 & 0.355 & 0.650 & 0.630 & 0.114 & 0.575 & 0.442 & \textbf{0.604} \\
\textit{OLMO 32B Instruct} & \textcolor{black}{0.000} & \textcolor{black}{0.411} & \textcolor{black}{0.688} & \textcolor{black}{0.621} & \textcolor{black}{0.273} & \textcolor{black}{0.561} & \textcolor{black}{0.432} & \textcolor{black}{0.593} \\
\textit{RedPajama-INCITE 7B Instruct} & \textcolor{black}{0.000} & \textcolor{black}{0.352} & \textcolor{black}{0.651} & \textcolor{black}{0.618} & \textcolor{black}{0.282} & \textcolor{black}{0.572} & \textcolor{black}{0.471} & \textcolor{black}{0.594} \\
\textit{MediPhi-PubMed} & \textcolor{black}{0.000} & \textcolor{black}{0.268} & \textcolor{black}{0.751} & \textcolor{black}{0.626} & \textcolor{black}{0.169} & \textcolor{black}{0.561} & \textcolor{black}{0.210} & \textcolor{black}{0.597} \\
\textbf{Prompting} &&&&&&&&\\
L3.3 70B 0-shot & 0.085 & 0.368 & 0.560 & -- & 0.236 & -- & 0.452 & -- \\
L3.3 70B 0-shot ordinal & 0.082 & 0.352 & 0.646 & -- & 0.289 & -- & \textbf{0.483} & -- \\
L3.3 70B few-shot & \textbf{0.095} & \textbf{0.380} & 0.652 & -- & \textbf{0.360} & -- & 0.452 & -- \\
L3.3 70B few-shot ordinal & 0.063 & 0.356 & 0.682 & -- & 0.313 & -- & 0.480 & -- \\
\textit{OLMO 32B 0-shot} & \textcolor{black}{0.018} & \textcolor{black}{0.333} & \textcolor{black}{0.652} & \textcolor{black}{--} & \textcolor{black}{0.322} & \textcolor{black}{--} & \textcolor{black}{0.430} & \textcolor{black}{--} \\
\textit{OLMO 32B 0-shot ordinal} & \textcolor{black}{0.006} & \textcolor{black}{0.323} & \textcolor{black}{\textbf{0.685}} & \textcolor{black}{--} & \textcolor{black}{0.325} & \textcolor{black}{--} & \textcolor{black}{0.443} & \textit{--} \\
\textit{OLMO 32B few-shot} & \textcolor{black}{0.014} & \textcolor{black}{0.316} & \textcolor{black}{0.630} & \textcolor{black}{--} & \textcolor{black}{0.219} & \textcolor{black}{--} & \textcolor{black}{0.423} & \textcolor{black}{--} \\
\textit{OLMO 32B few-shot ordinal} & \textcolor{black}{0.013} & \textcolor{black}{0.315} & \textcolor{black}{0.651} & \textcolor{black}{--} & \textcolor{black}{0.247} & \textcolor{black}{--} & \textcolor{black}{0.422} & \textcolor{black}{--} \\

\midrule
\textbf{Encoder-MLP head} &&&&&&&&\\
BERT & 0.264 & 0.550 & 0.871 & 0.501 & \textbf{0.441} & 0.498 & 0.614 & 0.501 \\
RoBERTa & 0.250 & 0.562 & 0.866 & 0.500 & 0.405 & 0.500 & 0.614 & 0.500 \\
DeBERTa-small & 0.262 & 0.570 & 0.871 & 0.601 & 0.382 & 0.498 & 0.617 & 0.556 \\
ModernBERT-base & 0.306 & \textbf{0.576} & 0.877 & 0.646 & 0.428 & 0.562 & 0.607 & 0.558 \\
ModernBERT-large & 0.277 & 0.568 & 0.874 & 0.598 & 0.391 & 0.539 & 0.602 & 0.553 \\
\textit{Bioclinical ModernBERT-base} & \textcolor{black}{0.264} & \textcolor{black}{0.559} & \textcolor{black}{\textbf{0.879}} & \textcolor{black}{\textbf{0.677}} & \textcolor{black}{0.395} & \textcolor{black}{\textbf{0.598}} & \textcolor{black}{\textbf{0.635}} & \textcolor{black}{\textbf{0.610}} \\
\textit{Bioclinical ModernBERT-large} & \textcolor{black}{\textbf{0.313}} & \textcolor{black}{0.562} & \textcolor{black}{0.868} & \textcolor{black}{0.671} & \textcolor{black}{0.399} & \textcolor{black}{0.491} & \textcolor{black}{0.618} & \textcolor{black}{0.607} \\
\midrule
\textbf{Encoder-masking-fine-tuned} &&&&&&&&\\
BERT & 0.166 & 0.432 & 0.690 & 0.648 & 0.323 & 0.554 & 0.482 & 0.600 \\
RoBERTa & 0.169 & 0.416 & 0.688 & 0.646 & 0.313 & 0.539 & 0.457 & 0.609 \\
DeBERTa-small & 0.175 & 0.435 & 0.695 & 0.671 & 0.325 & 0.579 & 0.492 & 0.617 \\
ModernBERT-base & \textbf{0.186} & 0.449 & \textbf{0.697} & 0.672 & 0.325 & 0.576 & \textbf{0.503} & 0.613 \\
ModernBERT-large & 0.177 & \textbf{0.460} & 0.683 & 0.663 & \textbf{0.342} & 0.615 & 0.496 & 0.632 \\
\textit{Bioclinical ModernBERT-base} & \textcolor{black}{0.166} & \textcolor{black}{0.450} & \textcolor{black}{0.692} & \textcolor{black}{\textbf{0.676}} & \textcolor{black}{0.336} & \textcolor{black}{0.604} & \textcolor{black}{0.499} & \textcolor{black}{\textbf{0.655}} \\
\textit{Bioclinical ModernBERT-large} & \textcolor{black}{0.165} & \textcolor{black}{0.422} & \textcolor{black}{0.696} & \textcolor{black}{0.652} & \textcolor{black}{0.325} & \textcolor{black}{\textbf{0.626}} & \textcolor{black}{0.425} & \textcolor{black}{0.603} \\
\midrule
\textbf{Encoder-masking-zeroshot} &&&&&&&&\\
BERT & 0.000 & 0.000 & 0.000 & 0.495 & 0.000 & 0.513 & 0.000 & 0.494 \\
RoBERTa & 0.049 & 0.207 & 0.375 & 0.500 & 0.137 & 0.500 & 0.318 & 0.500 \\
DeBERTa-small & 0.170 & 0.460 & 0.323 & 0.497 & 0.268 & \textbf{0.512} & 0.376 & \textbf{0.509} \\
ModernBERT-base & 0.169 & \textbf{0.468} & \textbf{0.804} & 0.496 & 0.358 & 0.501 & \textbf{0.511} & 0.499 \\
ModernBERT-large & 0.133 & 0.349 & 0.534 & 0.494 & 0.271 & 0.509 & 0.277 & 0.502 \\
\textit{Bioclinical ModernBERT-base} & \textcolor{black}{\textbf{0.246}} & \textcolor{black}{0.458} & \textcolor{black}{0.614} & \textcolor{black}{\textbf{0.498}} & \textcolor{black}{0.301} & \textcolor{black}{0.509} & \textcolor{black}{0.457} & \textcolor{black}{0.498} \\
\textit{Bioclinical ModernBERT-large} & \textcolor{black}{0.244} & \textcolor{black}{0.446} & \textcolor{black}{0.596} & \textcolor{black}{0.479} & \textcolor{black}{\textbf{0.364}} & \textcolor{black}{0.502} & \textcolor{black}{0.416} & \textcolor{black}{0.497} \\
\bottomrule
\end{tabular}
}
\caption{Forecasting performance (event occurrence: F1 and correct event ordering: concordance-index) of the ensuing $k=8$ events.  Models \emph{italicized} represent the open-source and medically fine-tuned models. All models are trained/fine-tuned on time-ordered annotations from \texttt{DeepSeek-R1}. \textbf{Bold} indicates best in that particular category amongst all models.}
\label{tab:DS_forecasting}
\end{table*}

\begin{table*}[!htbp]
\footnotesize
\centering
\resizebox{\linewidth}{!}{
\begin{tabular}{lcccccccc}
\toprule
& \multicolumn{4}{c}{\textit{T2S2} (\texttt{Llama-3.3})} & \multicolumn{2}{c}{\textit{sepsis-10}} & \multicolumn{2}{c}{\textit{sepsis-100}} \\
\cmidrule(lr){2-5} \cmidrule(lr){6-7} \cmidrule(lr){8-9}
& F1(1h) & F1(1d) & F1(1w) & c-index & F1(1d) & c-index & F1(1d) & c-index \\
\midrule
\textbf{LLM-MLP head} &&&&&&&&\\
DS-L3.3 70B & 0.140 & 0.563 & 0.811 & \textbf{0.624} & \textbf{0.397} & 0.585 & \textbf{0.629} & 0.598 \\
L3.3 70B & 0.265 & 0.466 & 0.807 & 0.619 & 0.160 & 0.567 & 0.466 & 0.597 \\ 
DS-L3.1 8B & 0.264 & \textbf{0.584} & \textbf{0.837} & \textbf{0.624} & 0.258 & 0.576 & 0.605 & 0.598 \\
L3.1 8B & \textbf{0.338} & 0.565 & 0.823 & 0.619 & 0.368 & 0.569 & 0.536 & 0.596 \\
\textit{OLMO 32B Instruct} & \textcolor{black}{0.190} & \textcolor{black}{0.315} & \textcolor{black}{0.764} & \textcolor{black}{0.604} & \textcolor{black}{0.238} & \textcolor{black}{0.551} & \textcolor{black}{0.315} & \textcolor{black}{0.596} \\
\textit{RedPajama-INCITE 7B Instruct} & \textcolor{black}{0.127} & \textcolor{black}{0.459} & \textcolor{black}{0.705} & \textcolor{black}{0.620} & \textcolor{black}{0.257} & \textcolor{black}{\textbf{0.586}} & \textcolor{black}{0.491} & \textcolor{black}{0.594} \\
\textit{MediPhi-PubMed} & \textcolor{black}{0.000} & \textcolor{black}{0.507} & \textcolor{black}{0.801} & \textcolor{black}{0.623} & \textcolor{black}{0.277} & \textcolor{black}{0.564} & \textcolor{black}{0.534} & \textcolor{black}{\textbf{0.600}} \\
\textbf{Prompting} &&&&&&&&\\
L3.3 70B 0-shot & 0.085 & 0.443 & 0.616 & -- & 0.236 & -- & 0.452 & -- \\
L3.3 70B 0-shot ordinal & 0.063 & 0.454 & 0.707 & -- & 0.289 & -- & \textbf{0.483} & -- \\
L3.3 70B few-shot & \textbf{0.099} & 0.457 & 0.703 & -- & \textbf{0.360} & -- & 0.452 & -- \\
L3.3 70B few-shot ordinal & 0.064 & \textbf{0.460} & 0.729 & -- & 0.313 & -- & 0.480 & -- \\
\textit{OLMO 32B 0-shot} & \textcolor{black}{0.036} & \textcolor{black}{0.417} & \textcolor{black}{0.708} & \textcolor{black}{--} & \textcolor{black}{0.254} & \textcolor{black}{--} & \textcolor{black}{0.473} & \textcolor{black}{--} \\
\textit{OLMO 32B 0-shot ordinal} & \textcolor{black}{0.034} & \textcolor{black}{0.457} & \textcolor{black}{\textbf{0.741}} & \textcolor{black}{--} & \textcolor{black}{0.167} & \textcolor{black}{--} & \textcolor{black}{0.427} & \textcolor{black}{--} \\
\textit{OLMO 32B few-shot} & \textcolor{black}{0.013} & \textcolor{black}{0.428} & \textcolor{black}{0.631} & \textcolor{black}{--} & \textcolor{black}{0.267} & \textcolor{black}{--} & \textcolor{black}{0.424} & \textcolor{black}{--} \\
\textit{OLMO 32B few-shot ordinal} & \textcolor{black}{0.046} & \textcolor{black}{0.413} & \textcolor{black}{0.647} & \textcolor{black}{--} & \textcolor{black}{0.300} & \textcolor{black}{--} & \textcolor{black}{0.424} & \textcolor{black}{--} \\

\midrule
\textbf{Encoder-MLP head} &&&&&&&&\\
BERT & 0.224 & 0.622 & 0.899 & 0.500 & 0.400 & 0.500 & 0.612 & 0.500 \\
RoBERTa & 0.180 & 0.608 & 0.900 & 0.500 & 0.364 & 0.500 & 0.597 & 0.500 \\
DeBERTa-small & 0.258 & 0.641 & 0.898 & 0.603 & 0.427 & 0.585 & 0.616 & 0.592 \\
ModernBERT-base & 0.257 & 0.645 & \textbf{0.903} & 0.594 & 0.431 & 0.590 & 0.626 & 0.565 \\
ModernBERT-large & \textbf{0.312} & 0.622 & 0.897 & 0.500 & 0.404 & 0.500 & 0.623 & 0.500 \\
\textit{Bioclinical ModernBERT-base} & \textcolor{black}{0.290} & \textcolor{black}{\textbf{0.653}} & \textcolor{black}{0.902} & \textcolor{black}{\textbf{0.650}} & \textcolor{black}{\textbf{0.449}} & \textcolor{black}{\textbf{0.618}} & \textcolor{black}{\textbf{0.662}} & \textcolor{black}{\textbf{0.627}} \\
\textit{Bioclinical ModernBERT-large} & \textcolor{black}{0.258} & \textcolor{black}{0.648} & \textcolor{black}{0.901} & \textcolor{black}{0.626} & \textcolor{black}{0.409} & \textcolor{black}{0.602} & \textcolor{black}{0.634} & \textcolor{black}{0.593} \\
\midrule
\textbf{Encoder-masking-fine-tuned} &&&&&&&&\\
BERT & 0.131 & 0.471 & 0.741 & 0.622 & 0.320 & 0.580 & 0.467 & 0.622 \\
RoBERTa & 0.129 & 0.484 & 0.659 & 0.625 & 0.329 & 0.587 & 0.472 & 0.614 \\
DeBERTa-small & 0.135 & 0.468 & 0.693 & 0.648 & 0.318 & 0.602 & 0.462 & 0.623 \\
ModernBERT-base & \textbf{0.139} & \textbf{0.489} & 0.700 & 0.632 & \textbf{0.334} & 0.595 & \textbf{0.478} & 0.612 \\
ModernBERT-large & 0.124 & 0.470 & 0.712 & 0.646 & 0.318 & 0.586 & 0.468 & 0.631 \\
\textit{Bioclinical ModernBERT-base} & \textcolor{black}{0.129} & \textcolor{black}{0.481} & \textcolor{black}{0.733} & \textcolor{black}{0.653} & \textcolor{black}{0.331} & \textcolor{black}{\textbf{0.615}} & \textcolor{black}{0.475} & \textcolor{black}{\textbf{0.658}} \\
\textit{Bioclinical ModernBERT-large} & \textcolor{black}{0.131} & \textcolor{black}{0.480} & \textcolor{black}{\textbf{0.840}} & \textcolor{black}{\textbf{0.658}} & \textcolor{black}{0.318} & \textcolor{black}{0.587} & \textcolor{black}{\textbf{0.478}} & \textcolor{black}{0.641} \\
\midrule
\textbf{Encoder-masking-zeroshot} &&&&&&&&\\
BERT & 0.000 & 0.000 & 0.000 & 0.504 & 0.000 & 0.515 & 0.000 & 0.506 \\
RoBERTa & 0.074 & 0.279 & 0.445 & 0.500 & 0.083 & 0.500 & 0.319 & 0.500 \\
DeBERTa-small & 0.008 & 0.439 & 0.839 & \textbf{0.562} & 0.296 & 0.518 & \textbf{0.515} & 0.501 \\
ModernBERT-base & 0.132 & \textbf{0.525} & \textbf{0.840} & 0.556 & \textbf{0.352} & 0.503 & 0.511 & 0.501 \\
ModernBERT-large & 0.109 & 0.413 & 0.622 & 0.499 & 0.326 & \textbf{0.521} & 0.339 & \textbf{0.518} \\
\textit{Bioclinical ModernBERT-base} & \textcolor{black}{\textbf{0.197}} & \textcolor{black}{0.460} & \textcolor{black}{0.533} & \textcolor{black}{0.499} & \textcolor{black}{0.299} & \textcolor{black}{0.520} & \textcolor{black}{0.465} & \textcolor{black}{0.499} \\
\textit{Bioclinical ModernBERT-large} & \textcolor{black}{0.194} & \textcolor{black}{0.451} & \textcolor{black}{0.525} & \textcolor{black}{0.496} & \textcolor{black}{0.342} & \textcolor{black}{0.489} & \textcolor{black}{0.433} & \textcolor{black}{0.501} \\
\bottomrule
\end{tabular}
}
\caption{Forecasting performance (event occurrence: F1 and correct event ordering: concordance-index) of the ensuing $k=8$ events. Models \emph{italicized} represent the open-source and medicallty fine-tuned models. \textbf{All models are trained/fine-tuned on time-ordered annotations from \texttt{Llama-3.3-70B}.}}
\label{tab:L33_forecasting}
\end{table*}

\clearpage
{\refstepcounter{section}
\noindent\Large\bfseries\thesection\quad Survival Performance on \texttt{DeepSeek-R1} Annotations for \textit{T2S2}, \textit{sepsis-10}, and \textit{sepsis-100}
\label{apd:full_survival}}
\vspace{1em}

In this section, we report the full concordance analyses for the \textit{sepsis-10} and \textit{sepsis-100} annotation sets in addition to \textit{T2S2} test set, presented in Table~\ref{tab:concordance_wide}. These results are the extended version of the \textit{T2S2}-only table (Table~\ref{tab:concordance_t2s2_only}) included in the main text.

\begin{table*}[!hbtp]
\centering
{
\begin{tabular}{lccccccccc}
\toprule
\textbf{Model} 
& \multicolumn{3}{c}{\textit{T2S2}} 
& \multicolumn{3}{c}{\textit{sepsis-10}} 
& \multicolumn{3}{c}{\textit{sepsis-100}} \\
\cmidrule(lr){2-4} \cmidrule(lr){5-7} \cmidrule(lr){8-10}

Time Threshold & 0h & 24h & 168h 
& 0h & 24h & 168h 
& 0h & 24h & 168h \\
\midrule
bert-base-uncased       & 0.60 & 0.61 & 0.53 & 1.00 & 1.00 & 1.00 & 0.58 & 0.62 & 0.58 \\
roberta-base            & 0.55 & 0.60 & 0.60 & 0.75 & 1.00 & 1.00 & 0.55 & 0.58 & 0.57 \\
deberta-v3-small        & 0.56 & 0.57 & 0.57 & 0.75 & 0.00 & 0.75 & 0.52 & 0.49 & 0.51 \\
ModernBERT-base         & 0.52 & 0.58 & 0.58 & 1.00 & 1.00 & 1.00 & 0.54 & 0.51 & 0.51 \\
ModernBERT-large        & 0.53 & 0.59 & 0.58 & 0.50 & 1.00 & 1.00 & 0.59 & 0.55 & 0.49 \\
\textcolor{black}{Bioclinical ModernBERT-base} & 0.57& 0.53& 0.53& 1.00& 0.75& 0.75& 0.63& 0.59& 0.56\\
\textcolor{black}{Bioclinical ModernBERT-large} & 0.54& 0.55 & 0.54 & 0.75 & 0.75 & 0.75& 0.61 & 0.57 & 0.49\\
DeepSeek-R1-Llama-70B   & \textbf{0.64} & \textbf{0.63} & 0.59 & 0.50 & 1.00 & 1.00 & 0.61 & 0.53 & 0.55 \\
Llama-3.3-70B-Instruct  & 0.62 & \textbf{0.63} & 0.58 & 0.50 & 1.00 & 0.75 & 0.64 & 0.57 & 0.53 \\
DeepSeek-R1-Llama-8B    & 0.60 & 0.58 & 0.58 & 1.00 & 1.00 & 1.00 & 0.53 & \textbf{0.66} & {0.65} \\
Llama-3.1-8B-Instruct   & 0.62 & 0.61 & \textbf{0.61} & 1.00 & 1.00 & 1.00 & 0.60 & 0.59 & 0.61 \\
\textcolor{black}{OLMO 32B Instruct} & 0.60& 0.62 & 0.59 & 0.75 & 0.75 & 0.75&\textbf{ 0.66} & 0.58 & 0.56\\
\textcolor{black}{RedPajama-INCITE 7B Instruct} & 0.56&  0.60 & \textbf{0.61} & 0.50 & 1.00 & 1.00& 0.61 & 0.53 & \textbf{0.76}\\
\textcolor{black}{MediPhi-PubMed} & 0.54 & 0.62 & 0.55 & 1.00 & 1.00 & 1.00& 0.62 & 0.58 & 0.61\\
\bottomrule
\end{tabular}
}
\caption{Survival analyses evaluated using time-dependent concordance index for each model across \texttt{DeepSeek-R1} test set annotations (\textit{T2S2}, \textit{sepsis-10}, \textit{sepsis-100}), evaluated at three observation time thresholds (0h, 24h, and 168h).
\vspace{-0.75em}
}
\label{tab:concordance_wide}
\end{table*}

\vspace{2em}

{\refstepcounter{section}
\noindent\Large\bfseries\thesection\quad Sensitivity Analysis of Time-Ordered versus Text-Ordered Training
\label{apd:sensitivity_analyses}}

\vspace{1em}

Table \ref{tab:dsr1q_l33_comparison} shows time-ordering generally improves concordance, particularly for \texttt{DeepSeek-R1} annotations, while F1 scores show mixed results. Text-ordering sometimes yields better external dataset performance with \texttt{Llama-3.3-70B}, suggesting that preserving the narrative sequence can support accurate classification even when it does not follow temporal order.  These results highlight a trade-off between temporal alignment and narrative coherence for different tasks.

\begin{table*}[ht]
\centering
{
\begin{tabular}{lcccccccc}
\toprule
& F1(1h) & F1(1d) & F1(1w) & c-index & F1(1d)-10 & c10 & F1(1d)-100 & c100 \\
\hline
\multicolumn{9}{c}{\textbf{\textit{T2S2} (\texttt{DeepSeek-R1})}} \\
\hline
\textbf{Text-ordered} & 0.290 & 0.569 & 0.876 & 0.659 & 0.361 & 0.598 & 0.615 & 0.609 \\
\textbf{Time-ordered} & 0.306 & 0.576 & 0.877 & 0.672 & 0.428 & 0.576 & 0.607 & 0.613 \\
\hline
\multicolumn{9}{c}{\textbf{\textit{T2S2} (\texttt{Llama-3.3-70B})}} \\
\hline
\textbf{Text-ordered} & 0.311 & 0.648 & 0.898 & 0.586 & 0.423 & 0.526 & 0.662 & 0.541 \\
\textbf{Time-ordered} & 0.257 & 0.645 & 0.903 & 0.632 & 0.382 & 0.595 & 0.616 & 0.612 \\
\bottomrule
\end{tabular}
}
\caption{Comparison of text-ordered and time-ordered training for \texttt{DeepSeek-R1} and \texttt{Llama 3.3-70B} annotations on F1 scores and c-index values. 
}
\label{tab:dsr1q_l33_comparison}
\end{table*}



    





\clearpage
{\refstepcounter{section}
\noindent\Large\bfseries\thesection\quad Masking History: \texttt{Llama-3.3-70B} Textual Time Series Annotations
\label{apd:masking_l33_dsr1}}

\vspace{1em}


Tables~\ref{tab:masking_effects_DS} and~\ref{tab:masking_effects_llama} report the effect of timestep dropout on classification and ranking performance. 
We evaluate performance across three datasets (\textit{T2S2}-\texttt{L3.3}, \textit{sepsis-10}, and \textit{sepsis-100}) using F1 scores for prediction of next \(k\) events happening within the next day and concordance index (c-index) for correctly ordering the next \(k\) events. 
As the proportion of masked history increases, we observe a general decline in F1 performance, particularly at higher masking rates (e.g., 90\%), suggesting that access to historical context is important for accurate forecasting. 
Interestingly, the concordance index remains relatively stable, indicating the model's robustness in ranking the next \(k\) events with partial input histories.

More specifically, we can see that F1 scores decrease monotonically with increasing TDR, with relatively mild degradation up to 30--60\% dropout and sharper declines thereafter. For example, in the \texttt{DeepSeek-R1} annotations (Table~\ref{tab:masking_effects_DS}), \textit{T2S2}/\textit{sepsis-10}/\textit{sepsis-100} drop from 0.576/0.428/0.607 at TDR = 0\% to 0.423/0.200/0.510 at TDR = 90\%, with \textit{sepsis-10} showing the largest relative decline. In contrast, concordance (c-index) remains stable across dropout levels (0.661--0.678), indicating that ranking-based evaluations are less sensitive to history truncation. These trends are consistent for \texttt{Llama-3.3} annotations (Table~\ref{tab:masking_effects_llama}), suggesting that ranking objectives capture temporal relationships that are more robust to missing historical information than binary classification accuracy of predicting whether an event happens within a time window.

\begin{table}[ht]
    \centering
    \begin{tabular}{lcccccc}
        \toprule
        \textbf{Timestep drop rate} & F1(1d) & c-index & F1(1d)-10 & c10 & F1(1d)-100 & c100 \\
        \hline
        0\%  & 0.576 & 0.672 & 0.428 & 0.576 & 0.607 & 0.613 \\
        15\% & 0.572 & 0.667 & 0.415 & 0.564 & 0.604 & 0.616 \\
        30\% & 0.548 & 0.672 & 0.357 & 0.542 & 0.591 & 0.619 \\
        60\% & 0.542 & 0.678 & 0.310 & 0.578 & 0.588 & 0.616 \\
        90\% & 0.423 & 0.661 & 0.200 & 0.561 & 0.510 & 0.614 \\
        \bottomrule
    \end{tabular}
    \caption{Effect of Randomly Masking History (\texttt{DeepSeek-R1} annotations) on F1 and Concordance. \textbf{Abbreviations}: F1(1d)-10/100: F1 (1 day) for \textit{sepsis-10} and \textit{sepsis-100} respectively. c10/100: c-index for \textit{sepsis-10} and \textit{sepsis-100} respectively.}
    \label{tab:masking_effects_DS}
\end{table}

\begin{table*}[!hbpt]
    \centering
    \resizebox{0.75\linewidth}{!}{
    \begin{tabular}{lcccccc}
        \toprule
        \textbf{Timestep drop rate} & F1(1d) & c-index & F1(1d)-10 & c10 & F1(1d)-100 & c100 \\
        \hline
        0\%  & 0.645 & 0.632 & 0.431 & 0.595 & 0.626 & 0.612 \\
        15\% & 0.628 & 0.635 & 0.422 & 0.592 & 0.624 & 0.610 \\
        30\% & 0.634 & 0.634 & 0.423 & 0.586 & 0.621 & 0.598 \\
        60\% & 0.604 & 0.634 & 0.391 & 0.577 & 0.605 & 0.613 \\
        90\% & 0.531 & 0.622 & 0.328 & 0.546 & 0.534 & 0.607 \\
        \bottomrule
    \end{tabular}
    }
    \caption{Effect of Randomly Masking History (\texttt{Llama-3.3-70B} annotations) on F1 and Concordance. \textbf{Abbreviations}: F1(1d)-10/100: F1 (1 day) for \textit{sepsis-10} and \textit{sepsis-100} respectively. c10/100: c-index for \textit{sepsis-10} and \textit{sepsis-100} respectively.}
    \label{tab:masking_effects_llama}
\end{table*}

\clearpage
\twocolumn
\section{Additional Insights into Forecasting Results}
\label{apd:additional_insights}

In this section, we provide some additional insights into the forecasting results. Specifically, we summarize two observations that clarify model behavior and apparent anomalies in the results.

\subsection{Why decoders fare better on survival but not short-horizon forecasting}
 Our findings align with adjacent literature on representation usage over different time horizons. \emph{Encoder} models, typically pooled via a \texttt{[CLS]} token, excel on short-horizon tasks because bidirectional attention aggregates local context effectively \citep{devlin2019bert,bai2018empirical}. In contrast, \emph{decoder} models trained autoregressively pass information forward through final token states, which is advantageous for long-range aggregation required by survival outcomes \citep{dai2019style,zhou2021informer}. In our setting, forecasting aligns with near-term prediction, whereas survival analysis emphasizes longer-range temporal dependencies; this mismatch naturally yields the observed performance split. While not a one-to-one architectural proof, the pattern is consistent with prior evidence on horizon-specific strengths \citep{wang2025large}.

\subsection{Interpreting very low $F_{1}$ scores and external validation drops}
 \textbf{Why some $F_{1}$ scores are near zero.} The very low F1 scores observed for certain models are not random failures but arise from two interacting factors. First, F1 is highly sensitive to threshold choice when the positive class is rare; small shifts in threshold calibration can yield 0.000 even when the model produces meaningful probabilities. Second, **architectural limitations** amplify this effect: **autoregressive decoders** tend to reproduce narrative rather than temporal order, and **instruction-tuned models** often default to generic or high-probability completions rather than selecting the correct event token—a behavior documented in prior analyses of instruction-following LLMs (\cite{liu2024lost, wei2021finetuned}).

These failures do not undermine model comparisons; rather, they highlight systematic architectural differences. **Encoder-based models**, trained with a **bidirectional masking objective** (\cite{devlin2019bert}), are better aligned with time-structured prediction tasks, as their contextual representation naturally integrates information from both past and future tokens. In contrast, **autoregressive decoders**, optimized for left-to-right generation (\cite{brown2020language}), are disadvantaged for temporal reasoning without explicit sequence restructuring or supervision. The contrast underscores that forecasting from clinical narratives is fundamentally different from standard text generation.
 
 \textbf{Why external validation is lower.} Performance drops on external subsets (e.g., \textit{sepsis-10} vs.\ \textit{sepsis-100}) reflect shifts in writing style and case mix, not random overfitting. Crucially, the \emph{relative ranking} of models remains stable across datasets, so comparative conclusions hold even when absolute scores vary.


\section{Open-Source Models for Avoiding Potential Data Leakage}
\label{apd:open_source}

We evaluated potential data leakage arising from large language models (LLMs) pretrained on web-scale corpora that may include PubMed Open Access (PMOA) case reports. Our safeguards target two avenues of leakage: (i) \emph{content overlap} with PMOA text, and (ii) \emph{task leakage} from using representations too close to raw prose. Below we summarize the controls embedded in our methods and the additional empirical evidence gathered for this concern.

\subsection{Task Design Minimizes Leakage}

\textbf{Temporal annotations—not raw text—drive all evaluations.}
We transform narratives into chronological lists of \emph{(event, time)} tuples and operate on these derived annotations (future-masked where applicable). Models are never scored on lexical similarity to the raw prose.

\textbf{Temporal reordering changes the learning problem.}
Events are reordered by extracted timestamps, breaking the original textual sequence seen during pretraining. This creates a distinct modeling challenge: correct temporal ordering and causal dependencies must be inferred from structured tuples rather than reproduced from prose. As shown in the main text, training/evaluating on time-ordered versus text-ordered sequences yields measurably different performance.

\textbf{Temporal metrics reduce any advantage from memorization.}
Evaluation emphasizes temporal reasoning—forecasting F1 at fixed horizons, pairwise concordance (c-index), and AULTC—rather than surface-form similarity. Memorizing text order can even \emph{hurt} time-alignment metrics because clinical narratives often interleave retrospective commentary where textual order $\neq$ chronological order.

\subsection{Open-Source, PubMed-Free Baselines}

\textbf{Decoder baseline.} 
\emph{RedPajama-INCITE-7B-Instruct} is pretrained on English CommonCrawl (67\%), C4 (15\%), GitHub (4.5\%), Wikipedia (4.5\%), Books (4.5\%), arXiv (2.5\%), and StackExchange (2\%), explicitly \emph{excluding} PubMed. In our experiments it performs comparably to decoders that may have seen PMOA during pretraining.

\textbf{Encoder baselines.} 
We include \emph{BERT} (BooksCorpus + Wikipedia), \emph{RoBERTa} (CommonCrawl News, OpenWebText, Stories, Wikipedia), and \emph{DeBERTa} (\textit{RoBERTa}-style corpora). None of these encoders include PubMed in their pretraining data.

\subsection{Post-Cutoff External Validation (Additional Evidence)}
To probe encoder leakage further, we evaluated \textit{BERT}, \textit{RoBERTa}, and \textit{DeBERTa} on 115 case reports published \emph{after} their pretraining cutoffs (post-2020). Results mirror submission trends without degradation, which is inconsistent with a memorization hypothesis.

\begin{table}[!htbp]
\centering
\footnotesize
\setlength{\tabcolsep}{6pt}
\renewcommand{\arraystretch}{1.1}
\begin{tabular}{lccc}
\hline
\textbf{Model} & \textbf{0h} & \textbf{24h} & \textbf{168h} \\
\hline
BERT     & 0.63 & 0.62 & 0.51 \\
RoBERTa  & 0.54 & 0.61 & 0.65 \\
DeBERTa  & 0.58 & 0.56 & 0.53 \\
\hline
\end{tabular}
\caption{Post-cutoff evaluation on 115 case reports (forecasting F1 at 0h/24h/168h).}
\label{tab:postcutoff_f1}
\end{table}

We also report competitive performance from \emph{OLMO-32B-Instruct} (fully documented training data) and parity from the PubMed-excluded \emph{RedPajama-INCITE-7B-Instruct} relative to decoders potentially exposed to PMOA, further indicating that prior PubMed exposure does not meaningfully advantage our temporally structured tasks.

The \textbf{key takeaways} are 
\begin{itemize}
    \item Using derived, time-ordered \emph{(event, time)} tuples and temporal metrics directly targets temporal reasoning and reduces susceptibility to content memorization.
    \item PubMed-free baselines and post-cutoff tests provide external checks
    \item Empirically, models without PubMed exposure achieve comparable performance, arguing against causal leakage into our forecasting, concordance, and AULTC evaluations.
\end{itemize}

\section{Generalizability Beyond Case Reports: Comparison with other Clinical Sources}
\label{apd:beyond_pmoa}

\subsection{Scope and Rationale for using Case Reports}
We acknowledge that published case reports differ from routine clinical documentation: they emphasize atypical presentations and are written for publication rather than point-of-care use, so they cannot substitute for validation on EHR notes. Our choice is methodological and pragmatic. Case reports are (i) publicly available, (ii) rich in temporally explicit clinical reasoning, and (iii) sufficiently structured to support extraction of textual time series. This makes them a practical, reproducible testbed for assessing whether forecasting models operate on temporally structured narratives without IRB barriers or restricted data access. We therefore use case reports to establish methodological feasibility and transparency, while recognizing that generalization to clinical operations must be demonstrated on EHR notes.

\subsection{Position Relative to Structured Clinical Scores}
Direct comparison to structured risk scores (e.g., SOFA, SAPS) is infeasible because required variables are absent from case reports. Our contribution is complementary to prior work that integrates text with structured data \citep{kline2022multimodal,wang2025large}: we show that free-text narratives, once transformed into timelines, carry prognostic signal usable for forecasting. Whereas \citet{kline2022multimodal} demonstrated that text can \emph{augment} structured models, our experiments isolate the incremental value of text when converted into time series, independent of structured covariates. In contexts where structured variables are available, these representations can be fused with established scores to assess additive value; here we focus on the text-only signal because the necessary structured inputs are not present in case reports.

\newpage
\section{Broader Societal Impact}
\label{apd:broader_impact}

This work advances equitable, scalable, and transparent medical AI by introducing a framework for forecasting clinical risk in sepsis from narrative documentation. Sepsis remains a leading cause of preventable mortality worldwide, with outcomes highly dependent on timely recognition and intervention. In many under‑resourced healthcare settings, structured EHR data and real‑time specialist input are unavailable, leaving unstructured narratives as the primary record of patient care. Our approach transforms such narratives into temporally structured representations, enabling systematic evaluation of forecasting methods for short‑ and long‑horizon event prediction, event ordering, and survival analysis. By releasing the dataset, extraction pipeline, and evaluation framework, we lower barriers for developing and benchmarking temporal forecasting models in sepsis, and promote reproducibility in LLM‑based clinical information extraction.

However, societal considerations are critical. LLM‑assisted extraction can introduce inaccuracies or hallucinations, and unvalidated forecasts risk amplifying bias or error. Although timelines are derived from non‑identifiable public case reports, their interpretation could still reinforce historical biases if not carefully contextualized. Forecasting results from case reports may also differ from those using routine clinical notes, limiting generalizability. Moreover, applying these forecasting models in high‑stakes clinical decision‑making without expert oversight poses safety risks. Responsible use requires transparent communication of dataset and model limitations, explicit uncertainty reporting, and safeguards to prevent inappropriate deployment in patient‑facing settings.


\end{document}